\documentclass[acmtog]{acmart}
\AtBeginDocument{%
  \providecommand\BibTeX{{%
    \normalfont B\kern-0.5em{\scshape i\kern-0.25em b}\kern-0.8em\TeX}}}

\acmSubmissionID{253}

\newcommand{\figref}[1]{Fig.~\ref{#1}}

\newcommand{\eqnref}[1]{Eq.~\ref{#1}}

\begin{document}

\title{Efficient Reflectance Capture with a Deep Gated Mixture-of-Experts}

\author{Xiaohe Ma}
\author{Yaxin Yu}
\author{Hongzhi Wu}
\affiliation{
  \institution{State Key Lab of CAD\&CG, Zhejiang University}
  \city{Hangzhou}
  \postcode{310058}
  \country{China}
}
\author{Kun Zhou}
\affiliation{
  \institution{State Key Lab of CAD\&CG, Zhejiang University and ZJU-FaceUnity Joint Lab of Intelligent Graphics}
  \city{Hangzhou}
  \postcode{310058}
  \country{China}
}

\renewcommand{\shortauthors}{Trovato and Tobin, et al.}

\begin{abstract}
We present a novel framework to efficiently acquire near-planar anisotropic reflectance in a pixel-independent fashion, using a deep gated mixture-of-experts. While existing work employs a unified network to handle all possible input, our network automatically learns to condition on the input for enhanced reconstruction. We train a gating module to select one out of a number of specialized decoders for reflectance reconstruction, based on photometric measurements, essentially trading generality for quality. A common, pre-trained latent transform module is also appended to each decoder, to offset the burden of the increased number of decoders. In addition, the illumination conditions during acquisition can be jointly optimized. The effectiveness of our framework is validated on a wide variety of challenging samples using a near-field lightstage. Compared with the state-of-the-art technique, our results are improved at the same input bandwidth, and our bandwidth can be reduced to about 1/3 for equal-quality results.
\end{abstract}



\begin{teaserfigure}
    \centering
    \includegraphics[width=\linewidth]{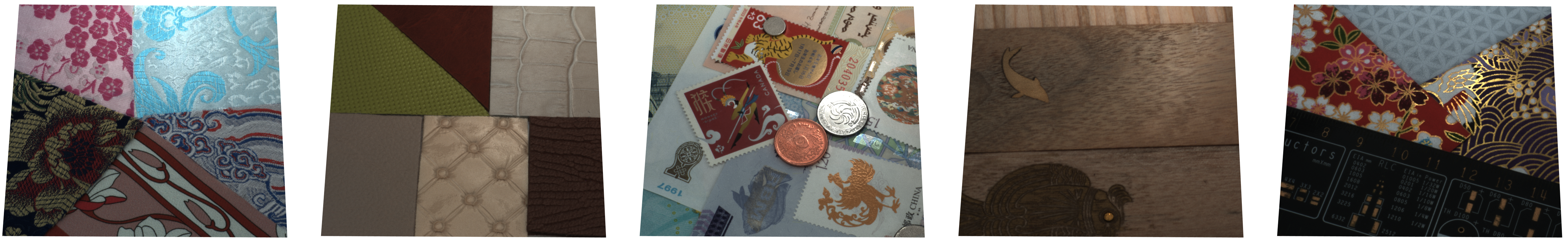}
    \caption{Rendering results of a variety of complex near-planar appearance, reconstructed using our neural network, with novel view and lighting conditions. Please also refer to the accompanying video for an animated sequence.}
  \label{fig:teaser}
\end{teaserfigure}

\thanks{*Corresponding author: Hongzhi Wu (\url{hwu@acm.org}).}
\maketitle
%
%
%
%

\section{Introduction}

High-quality digitization of physical material appearance is an important problem in computer graphics and vision, with a wide range of applications including visual effects, cultural heritage, e-commerce and computer games. The digital result, often represented as a 6D spatially-varying bidirectional reflectance distribution function (SVBRDF), can be rendered to faithfully reproduce the complex physical look that varies with location, lighting and view direction.

Directly capturing a general, near-planar reflectance sample can be performed with a spherical gantry, which exhaustively samples the combinations of all lighting and view directions~\cite{Dana:1999:BTF,Lawrence:2006:TREE}. This results in thousands or even millions of photographs, making it prohibitively expensive both in time and storage.

To improve the acquisition efficiency, one highly successful class of methods employ illumination multiplexing: instead of using a single source at a time, multiple lights are programmed simultaneously; the corresponding photometric measurements are then processed to produce the reflectance result in a pixel-independent manner. Representative work includes the lightstage~\cite{Ghosh:2009:GRADIENT,Tunwattanapong:2013:SH}, the linear light source reflectometry~\cite{Gardner:2003:LINEAR,Chen:2014:REFSCAN}, and setups with an LCD screen~\cite{Aittala:2013:FREQUENCY} or an LED array~\cite{Ma:2021:SCANNER}. Recently, neural acquisition techniques~\cite{Kang:2018:AUTO,Kang:2019:JOINT,Kang:2021:VIEWILLUMINATION} map both the physical acquisition and the computational reconstruction to a neural network, enabling the joint and automatic optimization of both processes. This leads to a substantially improved efficiency: 32 photographs for pixel-independent reconstruction of anisotropic reflectance from a single view~\cite{Kang:2018:AUTO}.

Our goal is to further push the limit of physical acquisition efficiency, as it is vital in applications like e-commerce, where the number of samples is large and the time budget on each sample is highly limited. We observe that state-of-the-art work is based on a unified neural network for all possible input, leading to a relatively lower processing efficiency, due to the potential interference effects. Inspired by the recent success of gated mixture-of-experts~\cite{Shazeer:2017:OUTRAGEOUSLY,Riquelme:2021:SVMOE}, we introduce a network that effectively \textbf{conditions on the input} for enhanced performance in reflectance acquisition.

In this paper, we propose a novel framework to adaptively learn to capture and reconstruct a near-planar SVBRDF. We automatically and jointly train a gating module to select one out of a number of specialized
decoders for optimal reflectance reconstruction, based on photometric measurements acquired with pre-optimized lighting patterns; each decoder is specifically tailored to efficiently handle a subset of possible reflectance only, essentially \textbf{trading generality for quality}. To alleviate the burden of the increasing number of decoders, we additionally pre-train a reflectance latent-space transform and 
simplify all decoders to output latent vectors only. Moreover, the illumination conditions during acquisition can be optimized in conjunction with the main network.

The effectiveness of our framework is demonstrated using an illumination multiplexing setup on 5 sets of challenging physical samples, with a wide variation in appearance. We improve the acquisition efficiency of anisotropic reflectance: for equal-bandwidth results, our reconstruction quality is above that of the state-of-the-art technique~\cite{Kang:2018:AUTO}, both qualitatively and quantitatively; for equal-quality results, we reduce the number of input photographs to 12 (corresponding to 6 seconds of acquisition time), in comparison with 32 as in~\cite{Kang:2018:AUTO}. Our results are validated against photographs, as well as rendered with novel lighting and view conditions. We hope that the key idea might also be helpful for boosting the performance in other neural acquisition tasks in the future.

\section{Related Work}

Below we mainly review existing work with active illumination, which is most related to this paper. For a comprehensive overview of reflectance acquisition, please refer to excellent recent surveys~\cite{Dong:2019:SURVEY,Weyrich:2009:SURVEY,Weinmann:2015:COURSE,GUARNERA:2016:BRDF}.

\subsection{Direct Sampling}
A straightforward approach to capture a general SVBRDF with high quality is to densely sample its 6D domain~\cite{Dana:1999:BTF,Lawrence:2006:TREE}. A spherical gantry captures photographs of the sample with a moving pair of a camera and a point light, effectively enumerating different combinations of the view and lighting directions. The acquisition process is prohibitively time consuming.

To improve the physical efficiency, various priors have been introduced to properly regularize the problem, while considerably reducing the number of measurements~\cite{Dong:2010:MANIFOLD,Wang:2008:ANISO,Lensch:2003:IMAGE,Marschner:1999:BRDF,Zickler:2005:SHARE,Nam:2018:FLASH,Hui:2017:ICCV}. Popular priors include restricting the reflectance to be isotropic, or assuming a strong coherence in the spatial domain (e.g., to express the material at each point as a linear combination of basis ones). The quality of reconstructed appearance might be limited, due to the lack of anisotropic reflections or intricate spatial details. In comparison, our approach does not rely on the aforementioned priors. Instead, we reconstruct complex anisotropic appearance in a pixel-independent fashion.

\subsection{Illumination Multiplexing}
Instead of using one light at a time, illumination-multiplexing-based approaches program the intensities of a number of sources simultaneously, substantially improving the acquisition efficiency and signal-to-noise-ratio. Traditional work first manually designs illumination conditions, captures corresponding responses of a material sample under such conditions, and finally recovers the reflectance properties from measurements (e.g., via an inverse look-up table)~\cite{Aittala:2013:FREQUENCY,Tunwattanapong:2013:SH,Nam:2016:MICRO,Ghosh:2009:GRADIENT,Gardner:2003:LINEAR,Chen:2014:REFSCAN,Ren:2011:POCKET}.

Recently, neural reflectance acquisition techniques map both the physical acquisition and computational processing to a single network, enabling the joint and automatic optimization of both the hardware and software~\cite{Kang:2018:AUTO,Kang:2019:JOINT,Ma:2021:SCANNER,Kang:2021:VIEWILLUMINATION}. Compared with traditional methods, this leads to nearly an order of magnitude increase in the acquisition efficiency. Our work is most similar to this line of work. Instead of employing a unified network, we adaptively process each input with a most suitable network, further boosting the sampling efficiency.

\subsection{Estimation from Highly Sparse Input}
Because of its practical value, SVBRDF estimation from a very small number of photographs, often with uncontrolled illumination, has received considerable attention in academia~\cite{Deschaintre:2018:SINGLE,Li:2017:SINGLE,Aittala:2015:TWO,Aittala:2016:NEURAL,Gao:2019:DIR,Guo::2021::TWOSTREAM}. This challenging problem is highly ill-posed, due to the huge gap in the amount of information between the limited input and the 6D output. Therefore, strong, hand-crafted or learning-based priors must be supplied to fill in this information gap. As a result, the estimation quality is affected: the spatial resolution of the output is usually limited; and general reflectance, such as anisotropic one, is not supported. Nevertheless, our idea of using a network that conditions on the input may be applied to work in this category as well to enhance their performance.

\section{Acquisition Setup}
\begin{figure}
     \centering
 	\includegraphics[width = \linewidth]{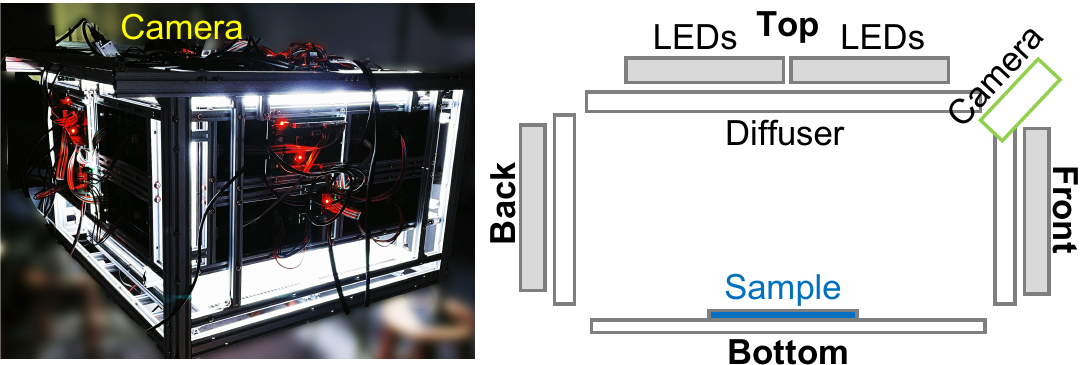}
     	\caption{The acquisition setup: a photograph (left) and a side view (right).}
   \label{fig:device}
\end{figure}

We build a hemicube-shaped, near-field lightstage to conduct physical experiments (\figref{fig:device}). Its size is about 70cm$\times$70cm$\times$40cm. We install a 24MP Basler a2A5328-15ucPRO vision camera to capture photographs of a near-planar material sample placed on the bottom plane of the device, from an angle of approximately $45^{\circ}$. The maximum size of the sample is 20cm$\times$20cm. There are 12,288 high-intensity RGB LEDs around the sample, attached with diffusers and mounted to the left, right, front, back and top sides of our setup. The LED pitch is 1cm, and the intensity is quantized with 8 bits and controlled using Pulse Width Modulation (PWM) with house-made circuits. We calibrate the intrinsic and extrinsic parameters of the camera, as well as the positions, orientations and the angular intensity distribution of each LED. In addition, vignetting is corrected with a flat field source, and color calibration is performed with an X-Rite ColorChecker Passport.

\section{Preliminaries}
Following~\cite{Kang:2018:AUTO}, we first list the relationship among the image measurement $B$ from a surface point $\mathbf{p}$, the reflectance $f$ and the intensity $I$ of each LED of the device. Below we focus on a single channel for brevity.
\begin{align}
B(I, \mathbf{x_{p}}, \mathbf{n_{p}}, \mathbf{t_{p}}) =  &\sum_{l}  I(l)  \int  \frac{1}{|| \mathbf{x_{l}} - \mathbf{x_{p}} ||^2} \Psi(\mathbf{x_{l}}, -\mathbf{\omega_{i}}) V(\mathbf{x_{l}}, \mathbf{x_{p}})  \nonumber \\
& f(\mathbf{\omega_{i}}'; \mathbf{\omega_{o}}', \mathbf{p}) (\mathbf{\omega_{i}} \cdot \mathbf{n_{p}})^{+} (-\mathbf{\omega_{i}} \cdot \mathbf{n_{l}})^{+} d\mathbf{x_{l}}.
\label{eq:render}
\end{align}
Here $l$ is the index of a planar light source, and $I(l)$ is its intensity in the range of [0, 1], the collection of which will be referred to as a lighting pattern in this paper. Moreover, $\mathbf{x_{p}}$/$\mathbf{n_{p}}$/$\mathbf{t_{p}}$ is the position/normal/tangent of $\mathbf{p}$, while $\mathbf{x_{l}}$/$\mathbf{n_{l}}$ is the position/normal of a point on the light whose index is $l$. We denote $\mathbf{\omega_{i}}$/$\mathbf{\omega_{o}}$ as the lighting/view direction, with $\mathbf{\omega_{i}} = \frac{\mathbf{x_{l}} - \mathbf{x_{p}}}{|| \mathbf{x_{l}} - \mathbf{x_{p}} ||}$. $\Psi(\mathbf{x_{l}}, \cdot)$ represents the angular distribution of the light intensity. $V$ is a binary visibility function between $\mathbf{x_{l}}$ and $\mathbf{x_{p}}$. The operator $( \cdot )^{+}$ computes the dot product between two vectors, and clamps a negative result to zero. $f(\cdot;  \mathbf{\omega_{o}}', \mathbf{p})$ is a 2D BRDF slice, which is a function of the lighting direction. We use the anisotropic GGX model~\cite{Walter:2007:GGX} to represent $f$, but other models can also be employed here.

Next, a lumitexel $m$ is defined as the collection of virtual measurements of the BRDF $f$ at a surface point $\mathbf{p}$, with one light on at a time~\cite{Lensch:2003:IMAGE,Kang:2018:AUTO}. It is a function of the light index $l$:
\begin{equation}
m(l; \mathbf{p}) = B(\{ I(l) = 1, \forall_{k \neq l } I(k) = 0 \}, \mathbf{p}).
\label{eq:lumitexel}
\end{equation}

\section{Overview}
We propose a deep gated mixture-of-experts network, to efficiently reconstruct the reflectance of a near-planar sample from single-view photographs under a set of pre-optimized lighting patterns. For each valid pixel location, the network first physically encodes the corresponding lumitexel as photometric measurements. Next, they are fed to the gating module to pick a suitable decoder, tailored for similar lumitexels. The decoder then transforms the same set of measurements to separately recover the diffuse/specular lumitexel. We fit a 4D BRDF along with a local frame to the decoded lumitexel at each pixel, which yields texture maps that represent the final 6D SVBRDF.
\begin{figure*}
\centering
	 \includegraphics[width=\linewidth]{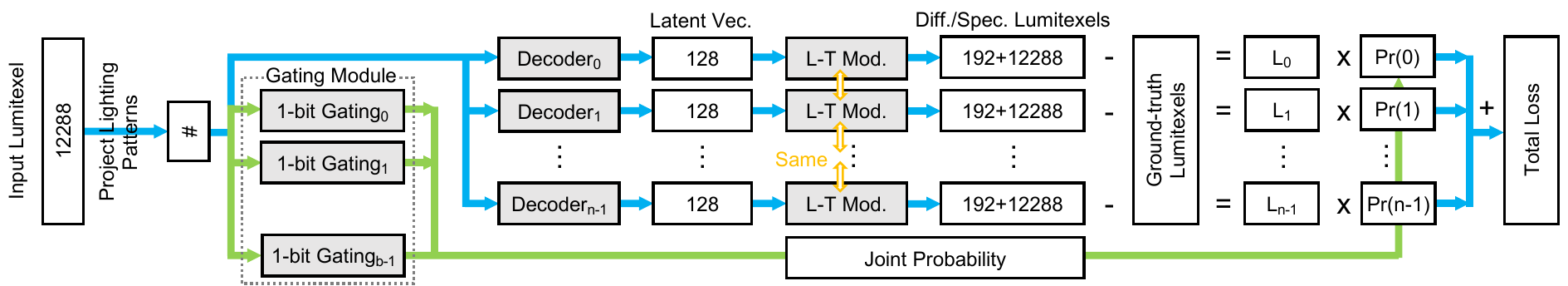}
    	\caption{Our network. It consists of a gating module, a total of $n$ specialized decoders and a latent-transform module. The gating module has $\log_2{n}$ single-bit gating subnets, the collection of their predictions determines a probability distribution over all decoders. The pre-trained latent-transform module (L-T Mod. in the figure) transforms the latent vector output from a decoder to a diffuse/specular lumitexel. The total loss is computed as the weighted average of the prediction loss of each decoder, using the aforementioned gating probability as weights.}
  \label{fig:pipeline}
\end{figure*}

\section{The Network}
Our goal is to introduce a differentiable framework that \textbf{automatically} learns to condition on the input for improved reflectance reconstruction quality. The idea is to first split the set of all possible input, and then process each subset separately. The reconstruction quality is expected to be improved, since each sub-space usually has a fractional size of the original space, and processing specialized to a sub-space can therefore trade generality for quality.

\subsection{Input/Output}
The input to our network is the set of \# physical measurements of a point on the material sample, captured with different pre-optimized lighting patterns. During training, the output is the diffuse/specular lumitexels reconstructed with different decoders. For testing, the output is the diffuse/specular lumitexel with a single selected decoder. We use \# to denote the number of measurements/lighting patterns. Note that similar to~\cite{Kang:2019:JOINT}, we only use a dimension of 192 to represent the diffuse lumitexel, due to its low-frequency nature.

\subsection{Architecture}
\begin{figure}
\centering
    \includegraphics[width = \linewidth]{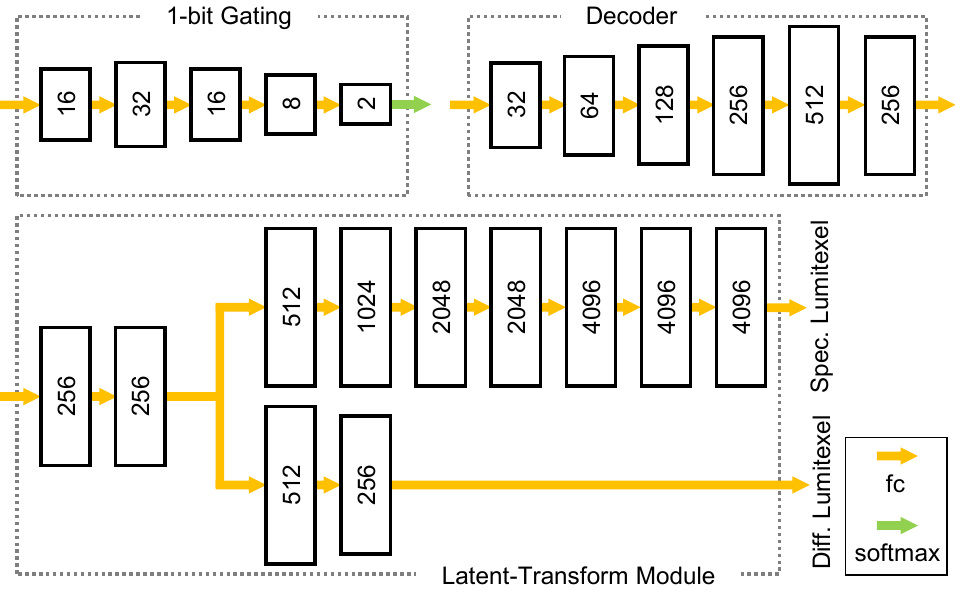}
	\caption{The network architecture of a 1-bit gating subnet, a decoder and the latent-transform module.Each fc layer before the last one is followed by a bn layer and then a leaky ReLU activation layer.}
  \label{fig:network}
\end{figure}

The main network consists of three parts: a gating module, a total of $n$ specialized decoders and a latent-transform module ($n=128$ in most experiments). Please refer to~\figref{fig:pipeline} for an overview and~\figref{fig:network} for architecture details. Each decoder has an index of a $\log_2n$-bit integer that starts from 0.

The gating module takes as input the photometric measurements at a pixel, and predicts a probability distribution over all decoders. It is made up of 5 fc layers followed by a softmax layer. Here the intuition is that more probabilities should be allocated to decoders that produce lower reconstruction losses for a given input, and vice versa. While a continuous probability distribution is predicted for differentiability, in experiments it often converges close to a 0-1 distribution at the end of training, essentially exploiting the best performing decoder. 

Specifically, the gating module consists of $\log_2n$ single-bit gating subnets. Each subnet takes as input the photometric measurements and outputs $g(k)$, the probability of the $k$-th bit of the index of the most suitable decoder being 1. Equivalently, for a decoder with an index of $a$, its chance of being picked by the gating can be computed as a joint probability:
\begin{equation}
    Pr(a) = \prod_{k=0}^{\log_2{n}-1} [a_k g(k) + (1-a_k) (1-g(k))],
\label{eq:gating}
\end{equation}
where $a_k$ denotes the $k$-th bit of $a$. Initially, we test a decision-tree-like, multi-level gating structure to compute the probability for each decoder based on a number of independent decisions; each is made based on a different subset of all measurements. This approach leads to less robust results, because individual decisions based on a subset of measurements are less reliable, compared to our current design that considers all measurements at once.

Next, each decoder takes as input the same photometric measurements and produces as output a latent code, which is further converted to a diffuse/specular lumitexel, by a pre-trained latent-transform module. One may employ decoders that directly generate lumitexels as output without the latent-transform module, given sufficient resources. That being said, we find it more efficient to exploit a latent space of all lumitexels as in the current design, as the intrinsic dimensionality of lumitexels is limited. This substantially reduces the size of each decoder, allowing us to train more of them for improved quality. Each decoder has the same structure with 7 fc layers.

Finally, the latent-transform module is pre-trained as part of an autoencoder, whose input is the physical lumitexel and the output is the corresponding diffuse/specular lumitexel. The dumbbell-shaped autoencoder has 17 fc layers. Its 128-D bottleneck corresponds to a latent vector of a lumitexel. After pre-training, we discard the part of the network prior to the bottleneck, and leave the remaining as the latent-transform module. Other work on the latent representation of 4D appearance may also be explored~\cite{Guo:2018:BRDFAnalysis,Hu:2020:DeepBRDF,rainer2020unified}.

Note that similar to previous work~\cite{Kang:2018:AUTO}, we link the lighting patterns during acquisition with the main network in a differentiable fashion, according to~\eqnref{eq:render}. This allows the joint optimization of the active illumination conditions, the gating and the decoders, for optimal reconstruction quality.

\subsection{Loss Function}
\label{sec:loss}
The loss function measures the squared difference between the predicted diffuse/specular lumitexels and their labels, for each decoder weighted by a probability determined by gating (\eqnref{eq:gating}):
\begin{align}
 L =  \sum_{a = 0}^{n-1} Pr(a) [ & \lambda_d \Sigma_l [m_d^a(l) - \tilde{m}_d(l)]^2 + \nonumber \\
 & \lambda_s \Sigma_l [\log(1+m_s^a(l)) - \log(1+\tilde{m}_s(l))]^2].
\label{eq:loss}
\end{align}
Here $m_d^a$/$m_s^a$ represents the diffuse/specular lumitexel predicted by the decoder with the index $a$, respectively. The corresponding ground-truths are denoted with a tilde. A $\log$ transform is performed to compress the high dynamic range in the specular reflectance. We use $\lambda_d = 1$ and $\lambda_s = 0.05$ in all experiments. Since the gating module affects $Pr(a)$, it gets optimized in conjunction with the decoders via back-propagation.

\subsection{Training}
Our network is implemented with PyTorch, and trained using the Adam optimizer with mini-batches of 50 and a momentum of 0.9. Xavier initialization is applied. Both the latent autoencoder and the main network are trained for 1 million iterations with a learning rate of $1\times10^{-4}$. Based on the GGX BRDF model and the calibration data of the device, we generate 200 million virtual lumitexels as training data (\eqnref{eq:render}), by randomly sampling the position, the shading frame, as well as BRDF parameters. Please refer to~\cite{Kang:2018:AUTO} for details.

For robustness in physical acquisition, we apply dropout regularization with a rate of 30\% to most layers, and perturb the synthetic measurements as well as sampled BRDF parameters with a multiplicative Gaussian noise ($\mu = 1$, $\sigma = 5\%$), similar to~\cite{Kang:2019:JOINT}. Moreover, we multiply a Gaussian noise ($\mu = 1$, $\sigma = 5\%$) to the input of the softmax layer in the gating module, to make it more resilient to potential measurement noise.

\subsection{Runtime}
\label{sec:post}
We first average the RGB channels of photometric measurements to a single gray-scale channel. The results are then sent to our network for gating computation, and the decoder with the highest $Pr$ is selected to produce a diffuse/specular lumitexel. Next, we nonlinearly fit a normal from the diffuse lumitexel, which serves as a good initialization for a subsequent fitting of the shading frame and roughness parameters from the specular lumitexel, using L-BFGS-B~\cite{Morales:2011:LBFGSB}. Finally, with the fixed shading frame and roughnesses, we compute the RGB diffuse/specular albedos, by solving non-negative linear least squares, constrained by the original photometric measurements, similar to~\cite{Ma:2021:SCANNER}.

\section{Results \& Discussions}
\label{sec:results}
In our setup, we capture the reflectance of 5 sets of near-planar physical samples (total number=29) with a wide variation in appearance. For a set of 12/32 lighting patterns, it takes 6/15 seconds in total to capture high-dynamic-range (HDR) images using exposure bracketing. Similar to~\cite{Kang:2019:JOINT}, a lighting pattern that contains both positive and negative weights is split into two for physical realization: one containing all positive weights with others set to zero, and the other with all negative weights sign-flipped and others set to zero. Throughout this paper, we report the number of physically realized lighting patterns for consistency.
\begin{figure}
    \centering
    \includegraphics[width=0.24\linewidth]{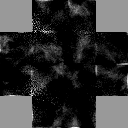}
    \includegraphics[width=0.24\linewidth]{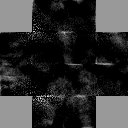}
    \includegraphics[width=0.24\linewidth]{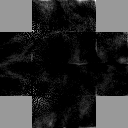}
    \includegraphics[width=0.24\linewidth]{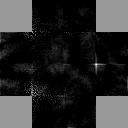}

    \includegraphics[width=0.24\linewidth]{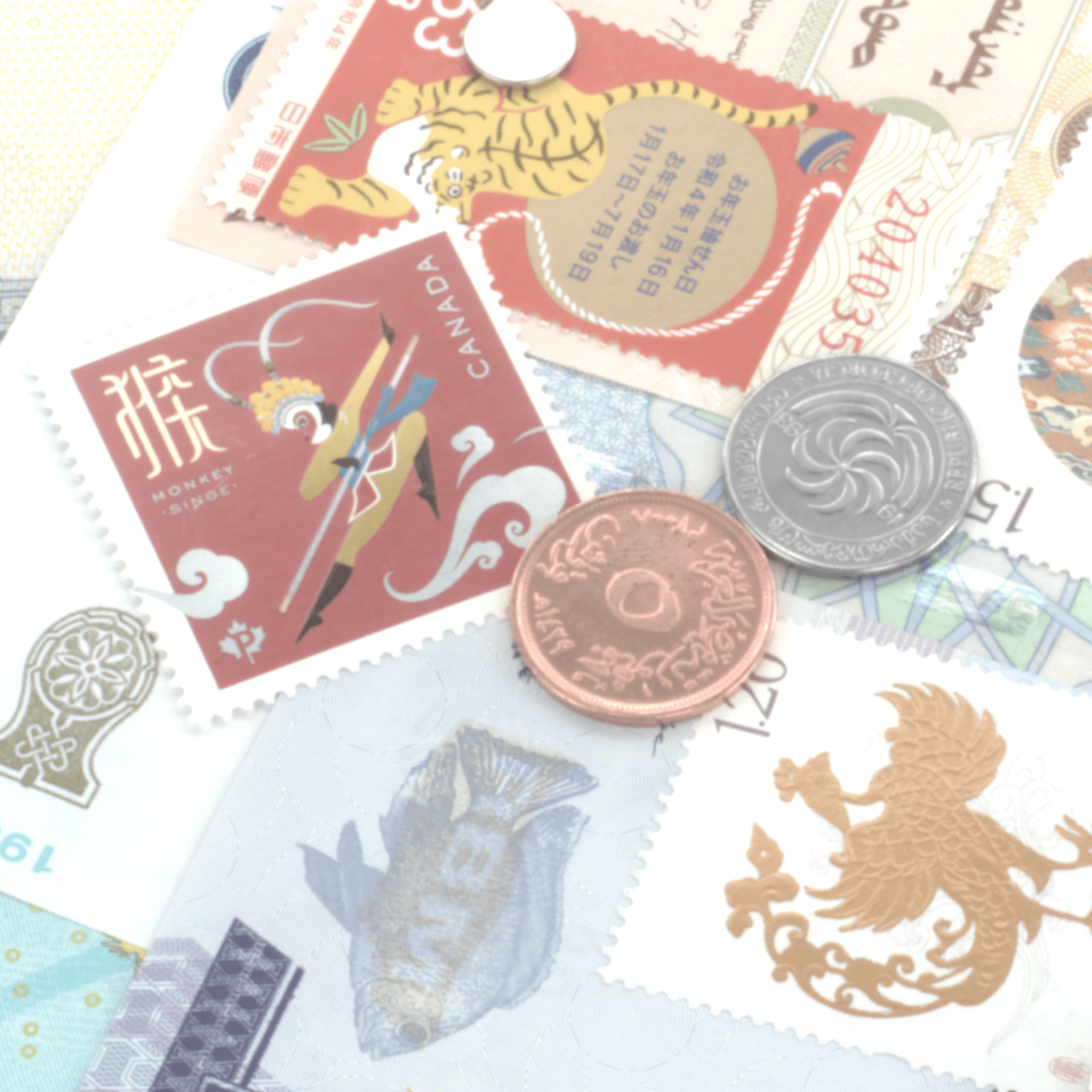}
    \includegraphics[width=0.24\linewidth]{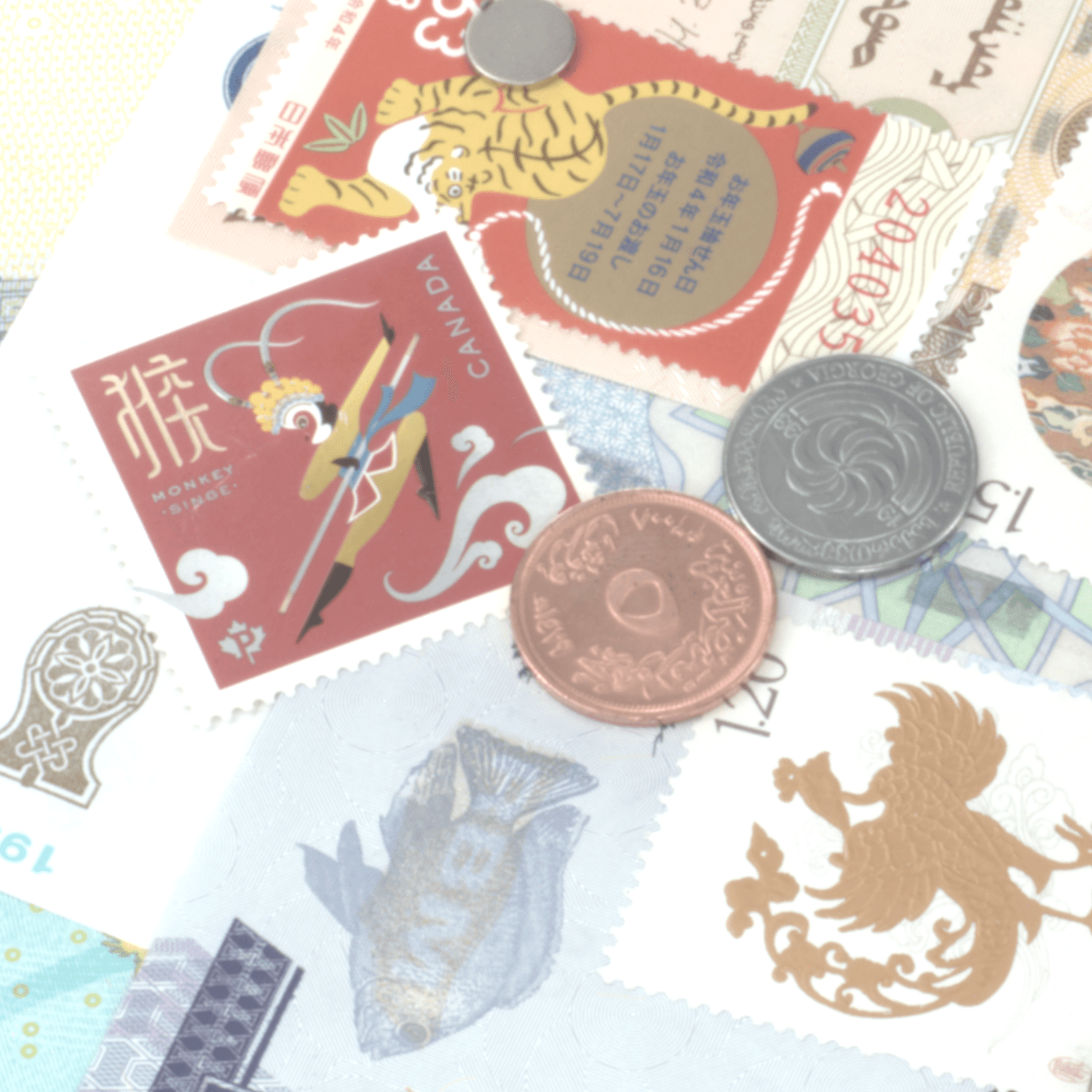}
    \includegraphics[width=0.24\linewidth]{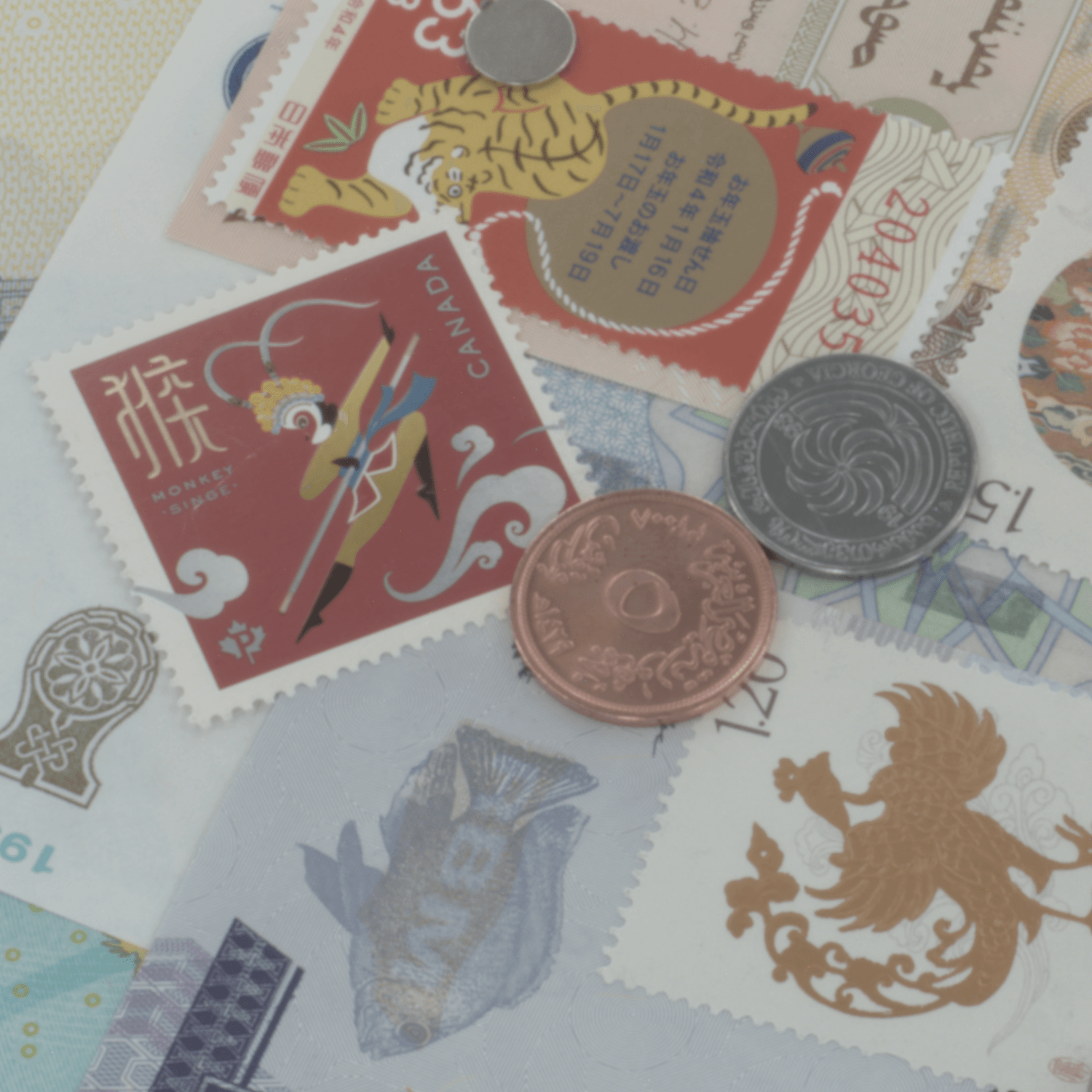}
    \includegraphics[width=0.24\linewidth]{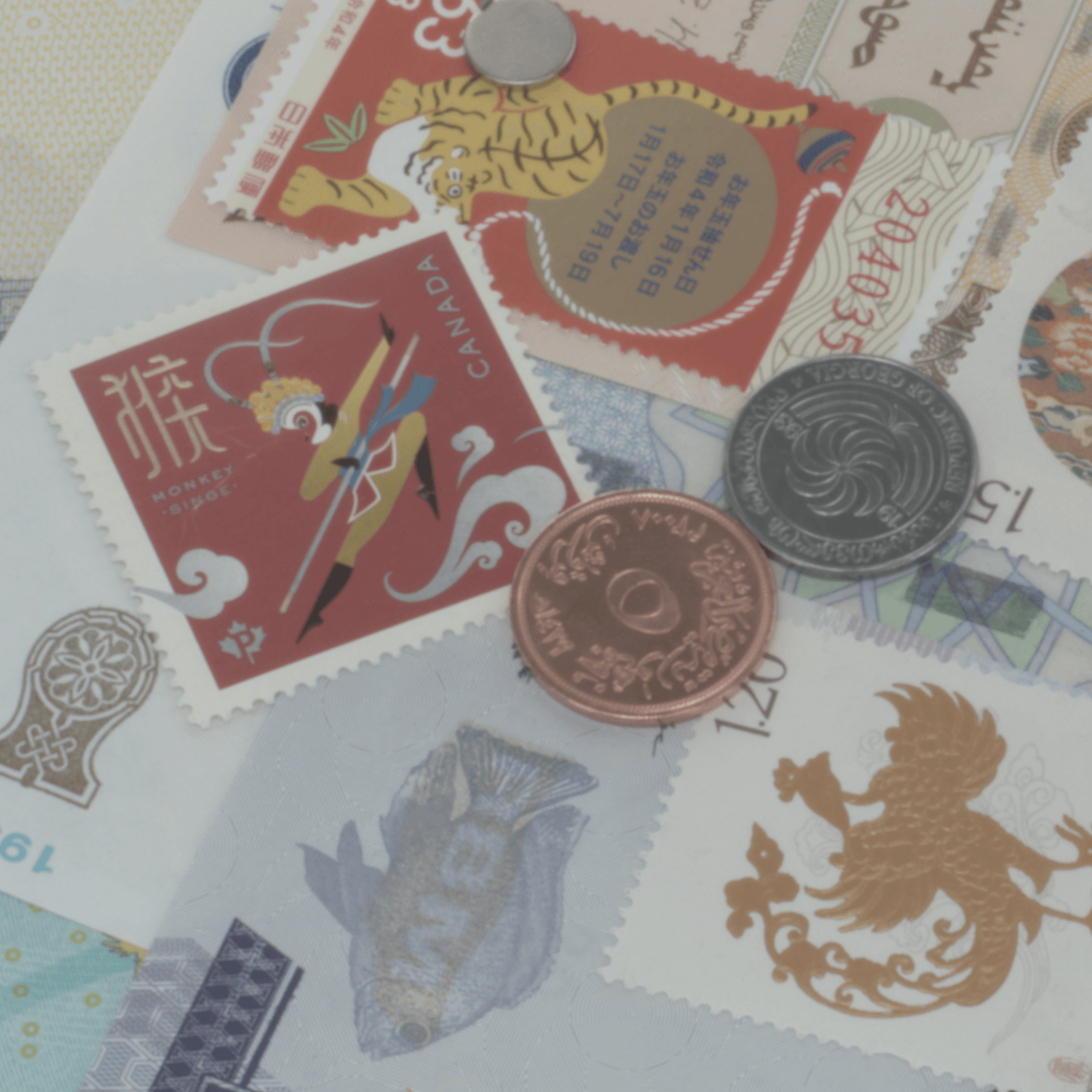}

    \includegraphics[width=0.24\linewidth]{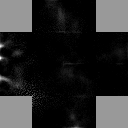}
    \includegraphics[width=0.24\linewidth]{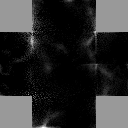}
    \includegraphics[width=0.24\linewidth]{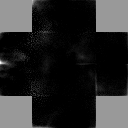}
    \includegraphics[width=0.24\linewidth]{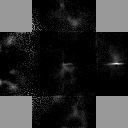}

    \includegraphics[width=0.24\linewidth]{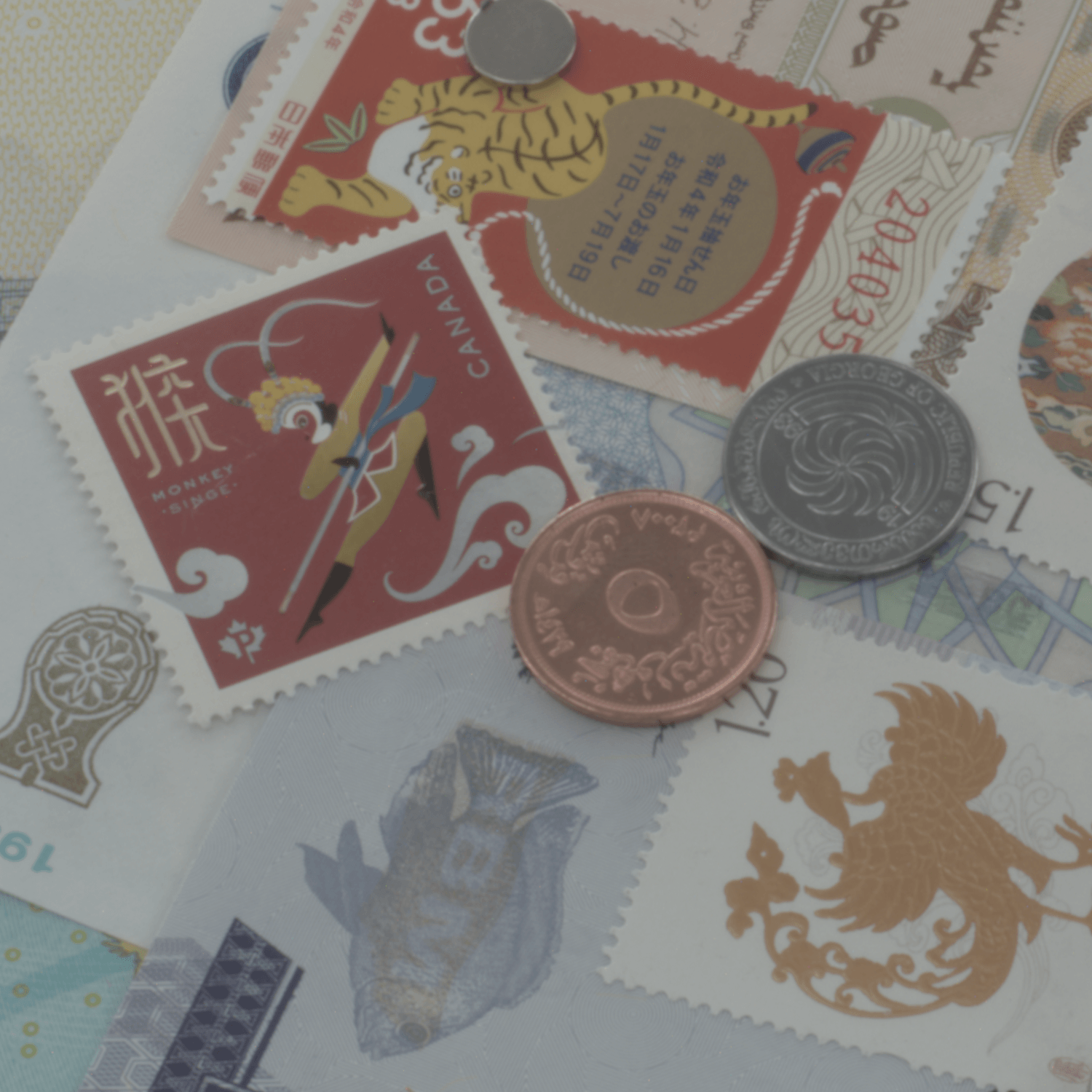}
    \includegraphics[width=0.24\linewidth]{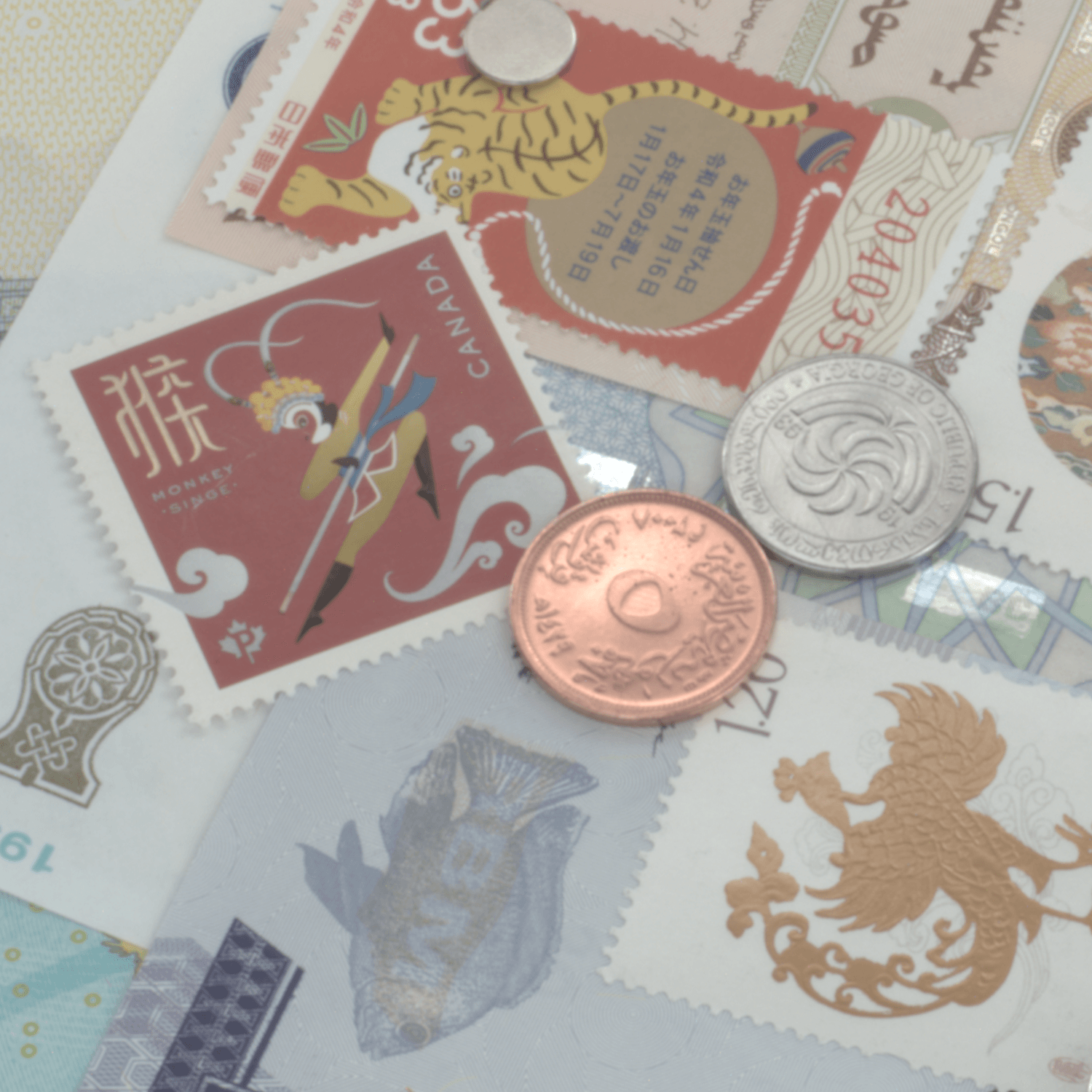}
    \includegraphics[width=0.24\linewidth]{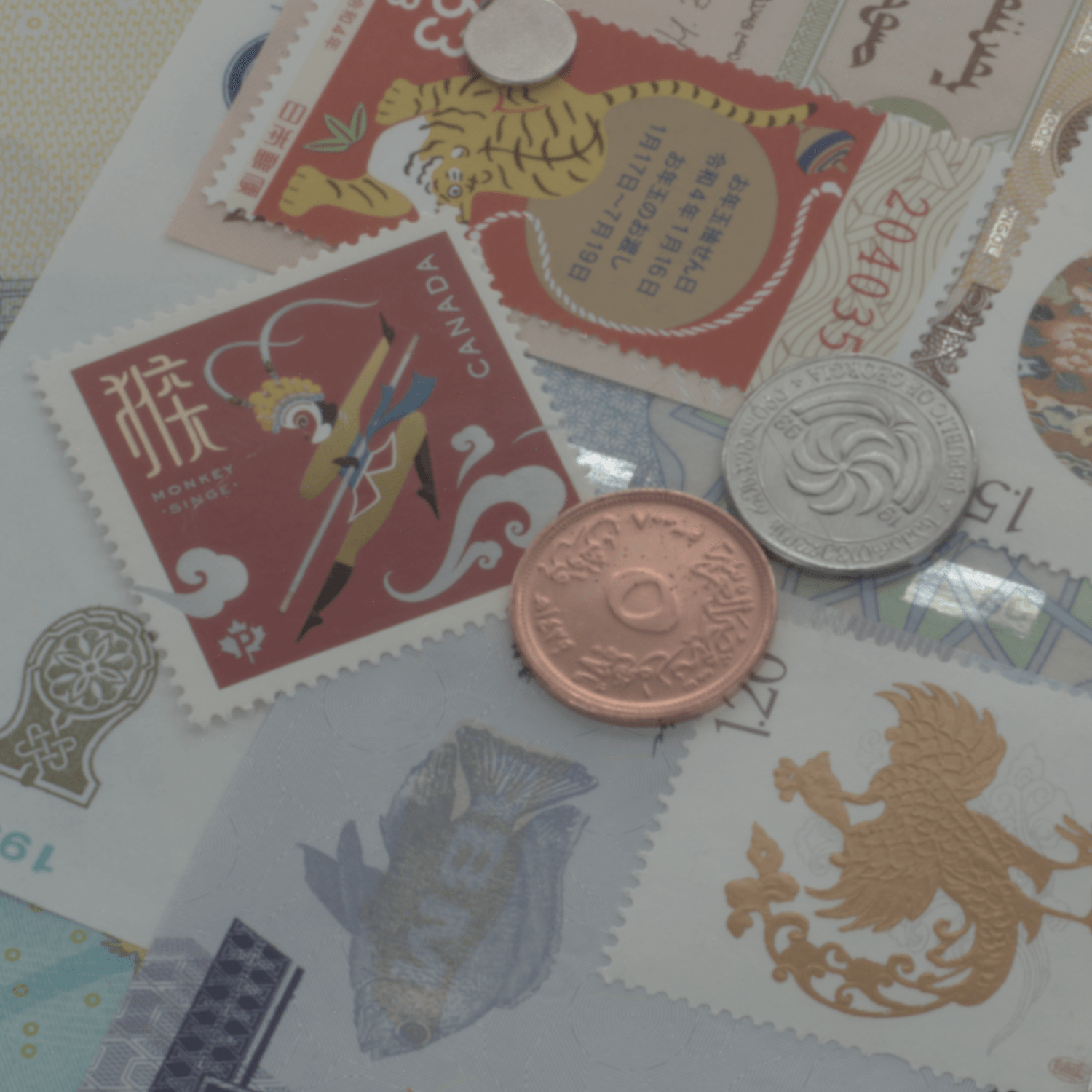}
    \includegraphics[width=0.24\linewidth]{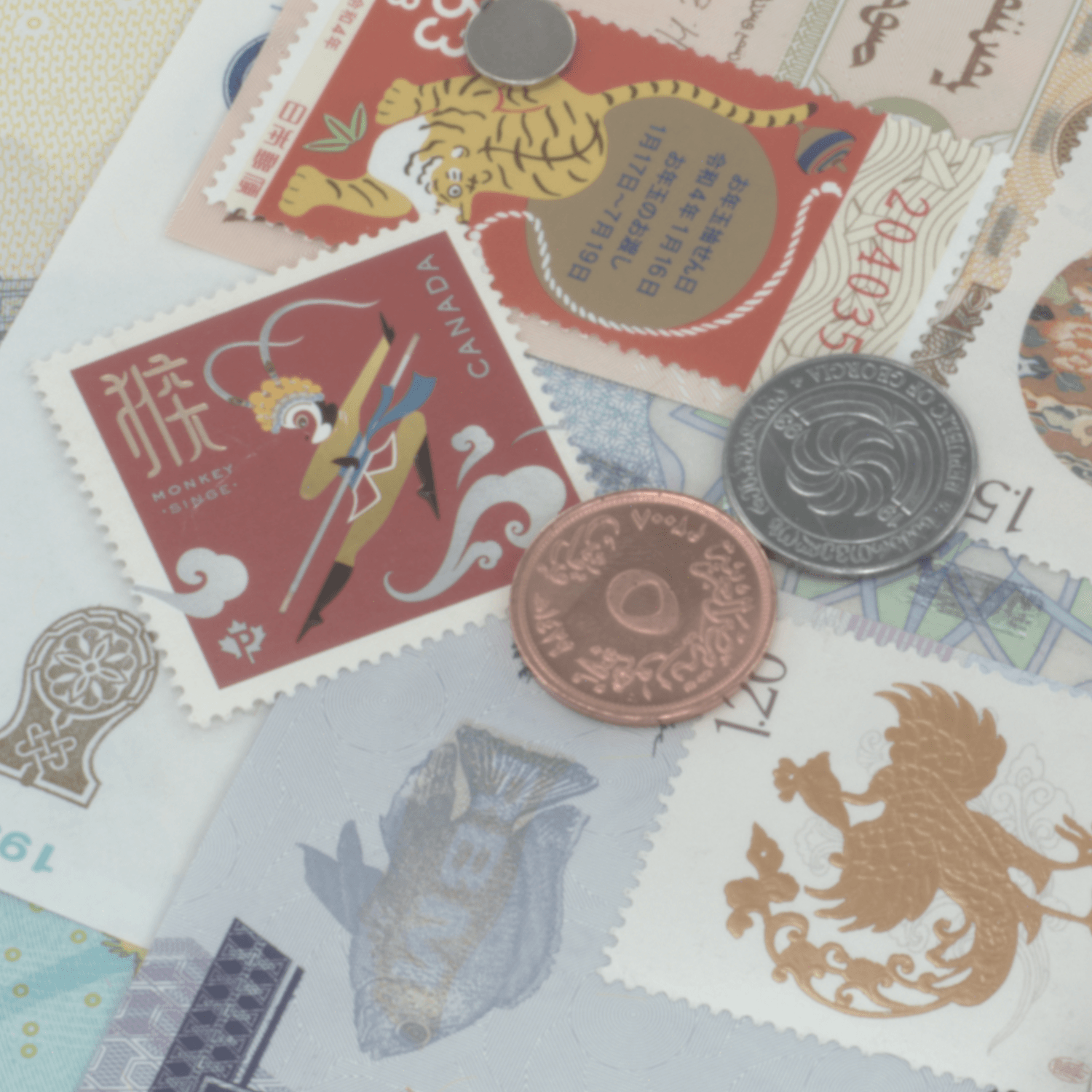}

   \caption{Visualization of different lighting patterns. From the top row to bottom: our patterns, photographs of a sample under our patterns, patterns trained with~\cite{Kang:2018:AUTO} and corresponding photographs. The lighting pattern is parameterized on a cross, by unfolding all side faces to the top plane. Note that only a subset of all patterns are shown due to limited space.} 
   \label{fig:compare_pattern}
\end{figure}

All computation is done on a workstation with dual Intel Xeon 4210 CPUs, 256GB DDR4 memory and 4 NVIDIA GeForce RTX 3090 GPUs. It takes on average 72 hours to train our network for 1 million iterations. The latent autoencoder takes 60 hours to pre-train. At runtime, it takes 5 minutes for our network to decode 1 million pairs of diffuse/specular lumitexels from measurements, and 1.5 hours for the subsequent GGX parameter fitting. The timing is comparable to existing work~\cite{Kang:2018:AUTO}. We use a spatial resolution of $1024^2$ to store all GGX parameters.

\begin{figure*}
            \centering
            \includegraphics[width=1.35in]{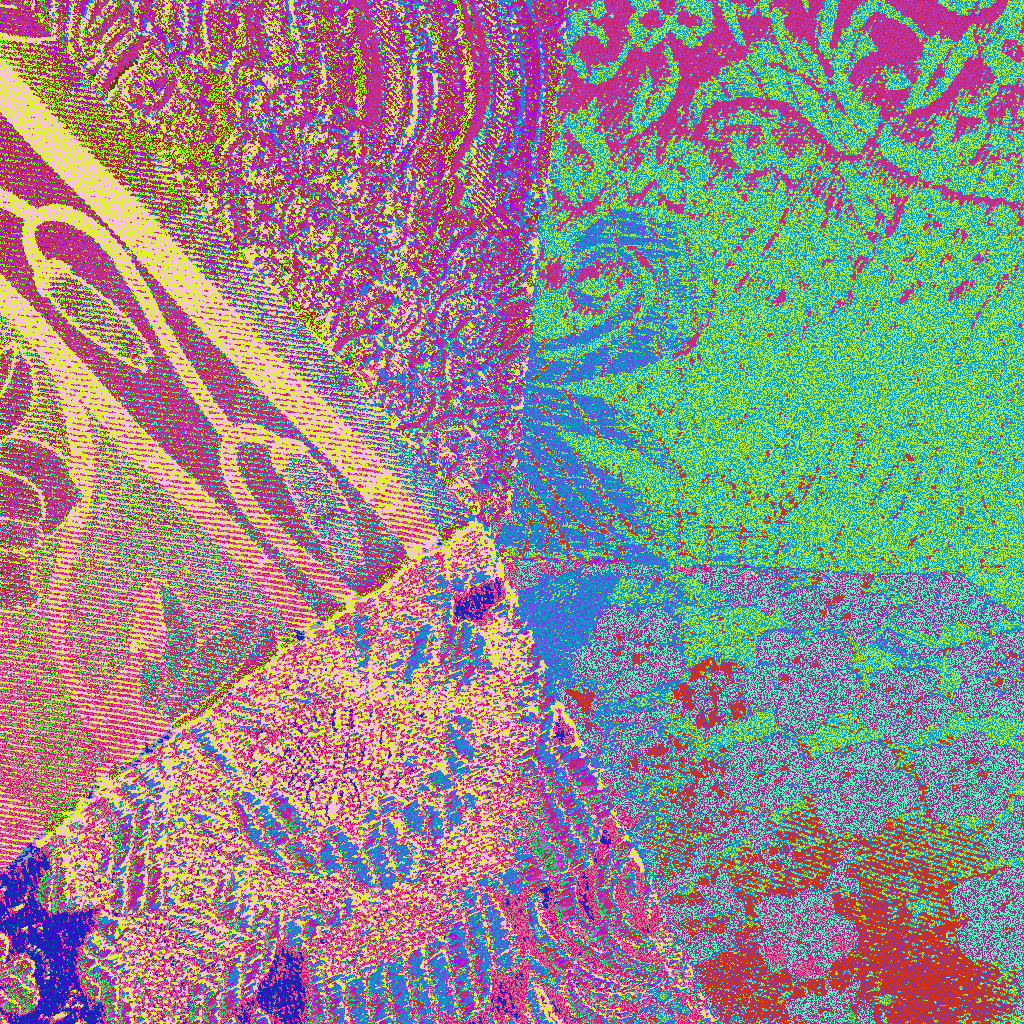}
            \includegraphics[width=1.35in]{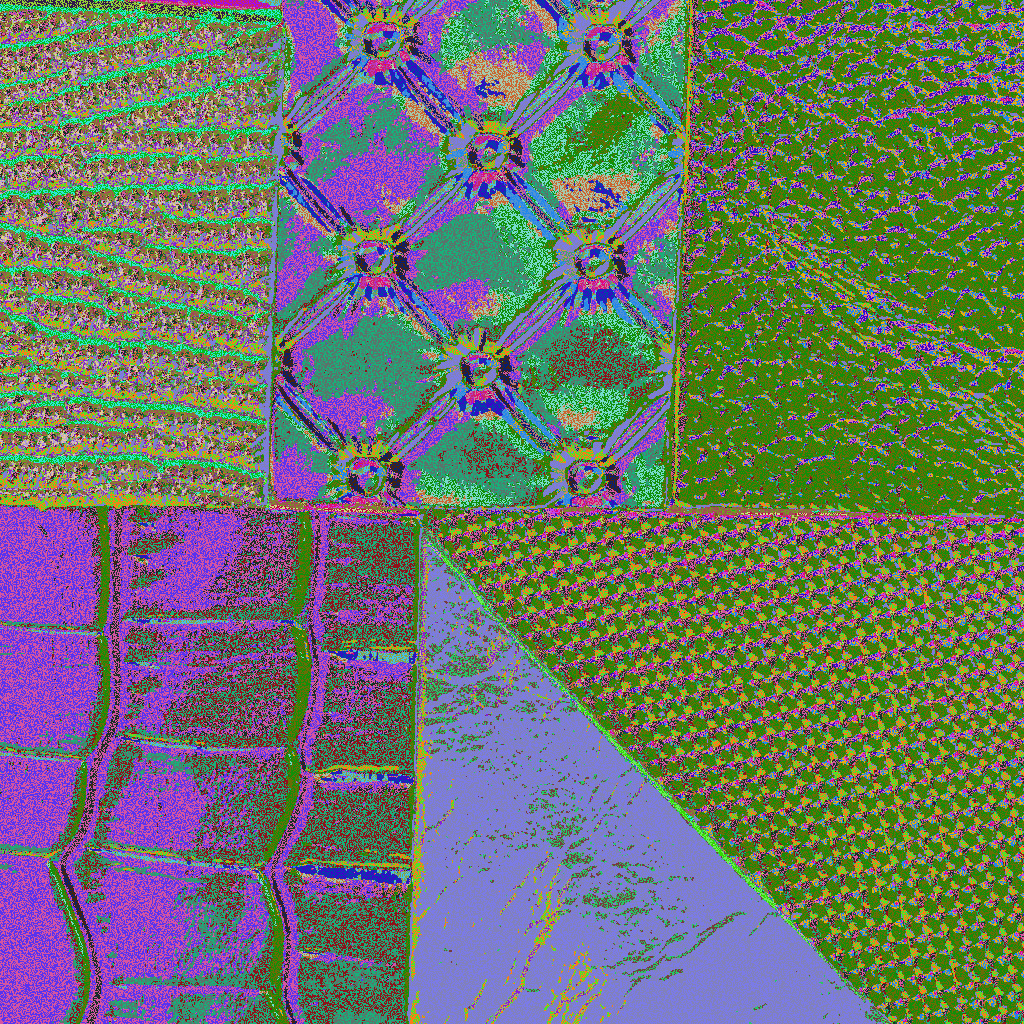}
            \includegraphics[width=1.35in]{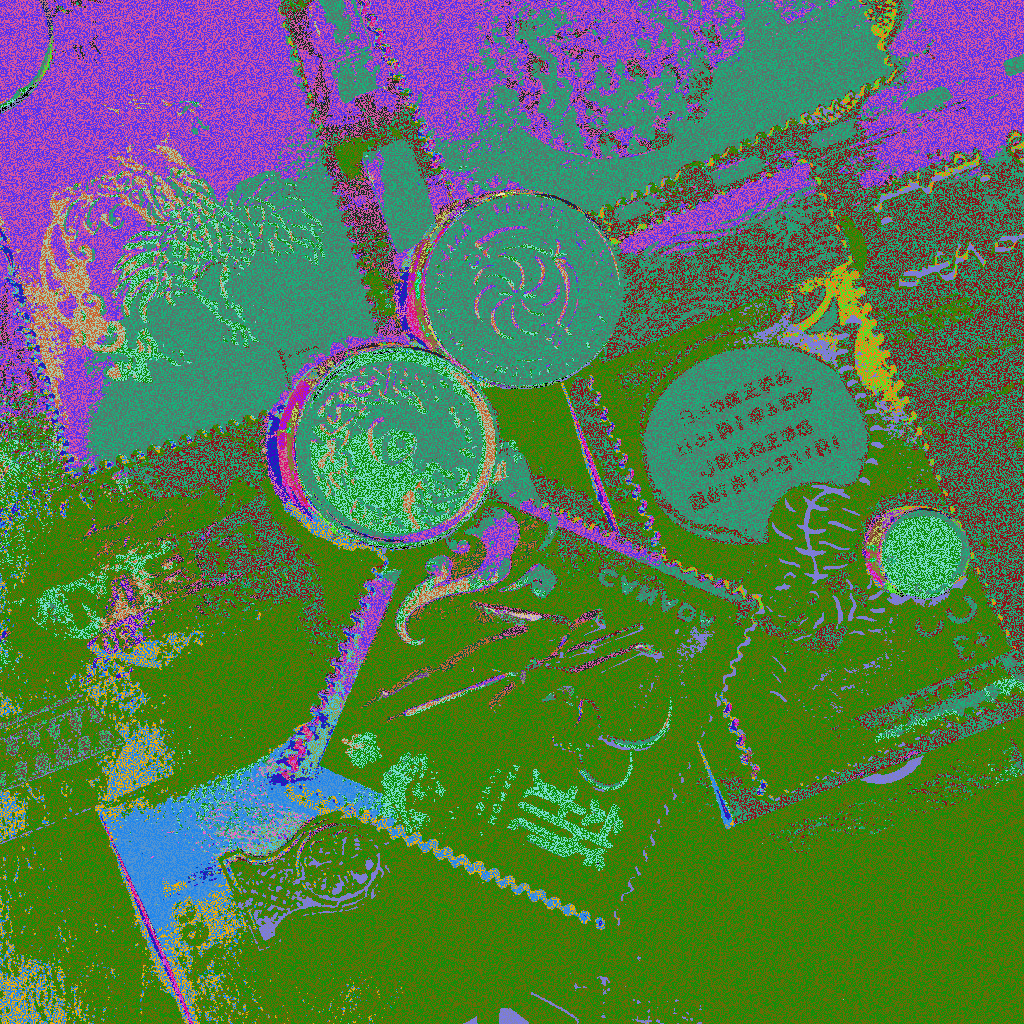}
            \includegraphics[width=1.35in]{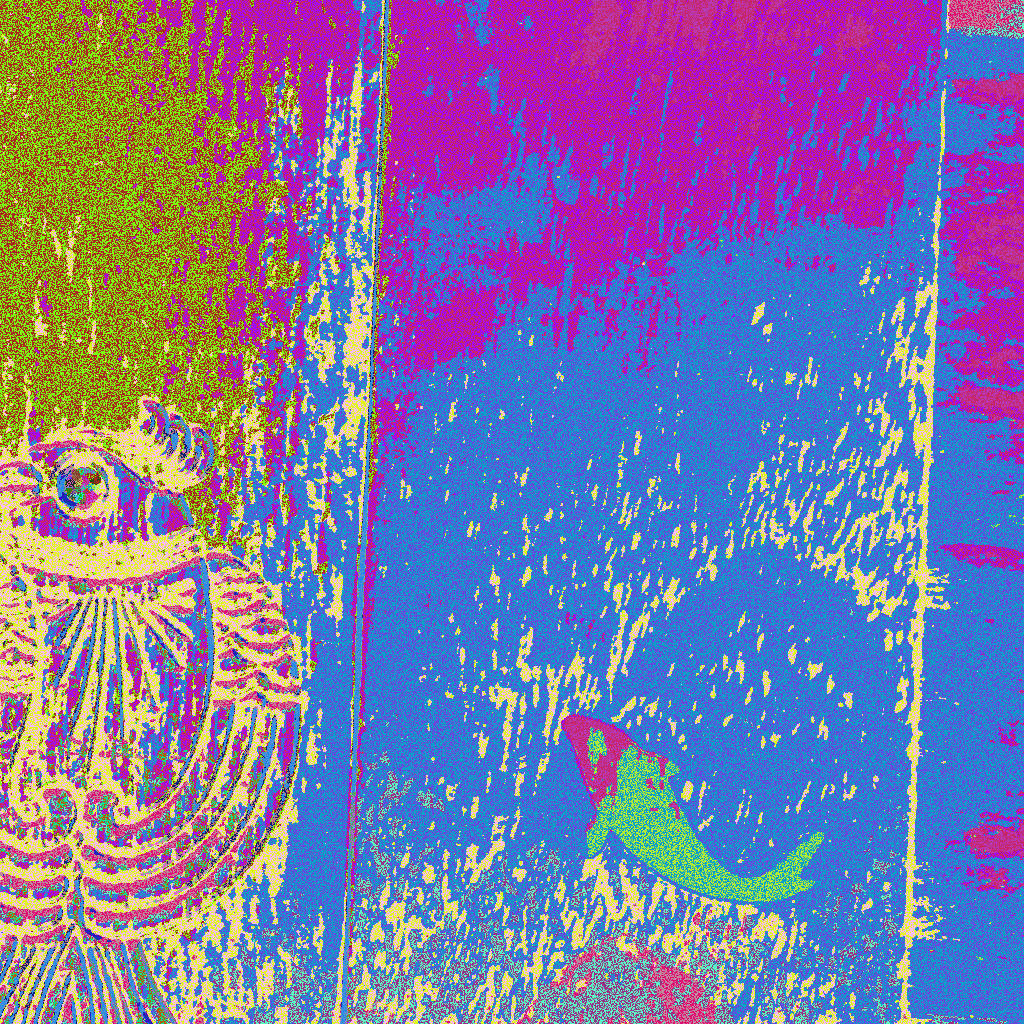}
            \includegraphics[width=1.35in]{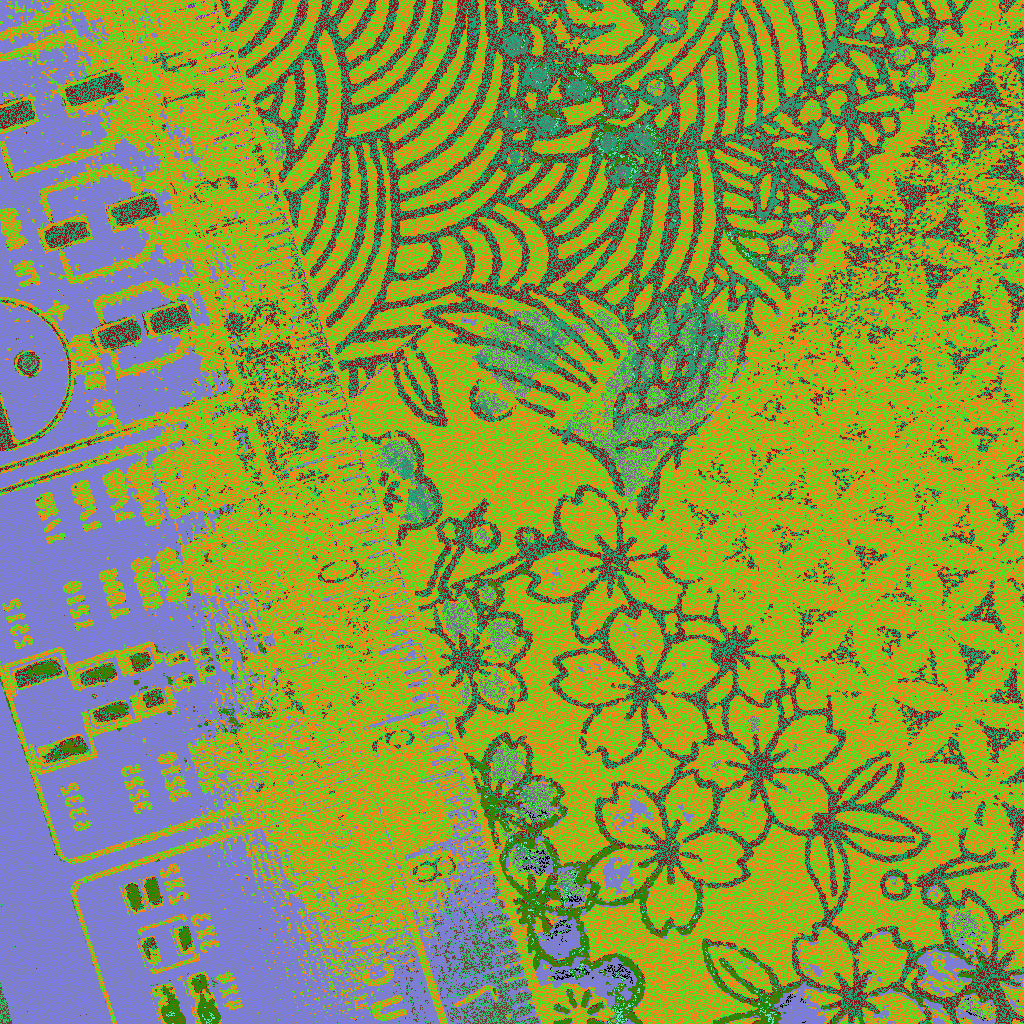}
  \caption{Visualization of the gating results. Each pixel shows the color-coded index of the decoder with the maximum predicted probability.}
    \label{fig:gating}
\end{figure*}

\figref{fig:compare_pattern} visualizes our lighting patterns along with those trained using~\cite{Kang:2018:AUTO}. One can observe that our patterns exhibit more high-frequency details. The captured photographs of a physical sample set can be found in the same figure. Moreover, the gating result at each pixel (i.e., the index of the decoder with the highest predicted probability) is visualized in~\figref{fig:gating}. Our gating module automatically learns to cluster pixels with similar high-dimensional appearance for efficient processing.

\begin{figure}
    \centering
    \includegraphics[width = 0.52in]{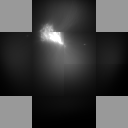}
    \includegraphics[width = 0.52in]{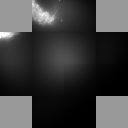}
    \includegraphics[width = 0.52in]{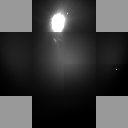}
    \includegraphics[width = 0.52in]{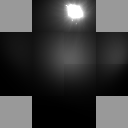}
    \includegraphics[width = 0.52in]{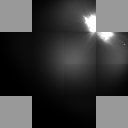}
    \includegraphics[width = 0.52in]{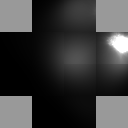}
    
    \includegraphics[width = 0.52in]{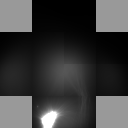}
    \includegraphics[width = 0.52in]{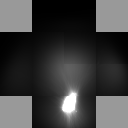}
    \includegraphics[width = 0.52in]{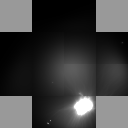}
    \includegraphics[width = 0.52in]{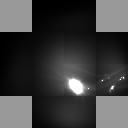}
    \includegraphics[width = 0.52in]{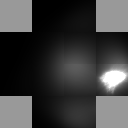}
    \includegraphics[width = 0.52in]{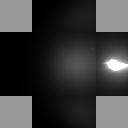}
    
    \includegraphics[width = 0.52in]{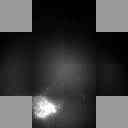}
    \includegraphics[width = 0.52in]{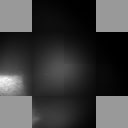}
    \includegraphics[width = 0.52in]{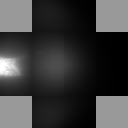}
    \includegraphics[width = 0.52in]{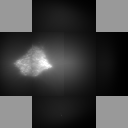}
    \includegraphics[width = 0.52in]{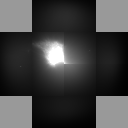}
    \includegraphics[width = 0.52in]{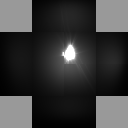}

    \includegraphics[width = 0.52in]{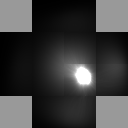}
    \includegraphics[width = 0.52in]{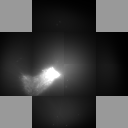}
    \includegraphics[width = 0.52in]{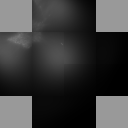}
    \includegraphics[width = 0.52in]{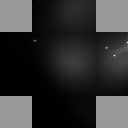}
    \includegraphics[width = 0.52in]{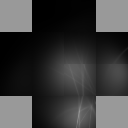}
    \includegraphics[width = 0.52in]{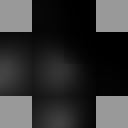}
    \includegraphics[width = \linewidth]{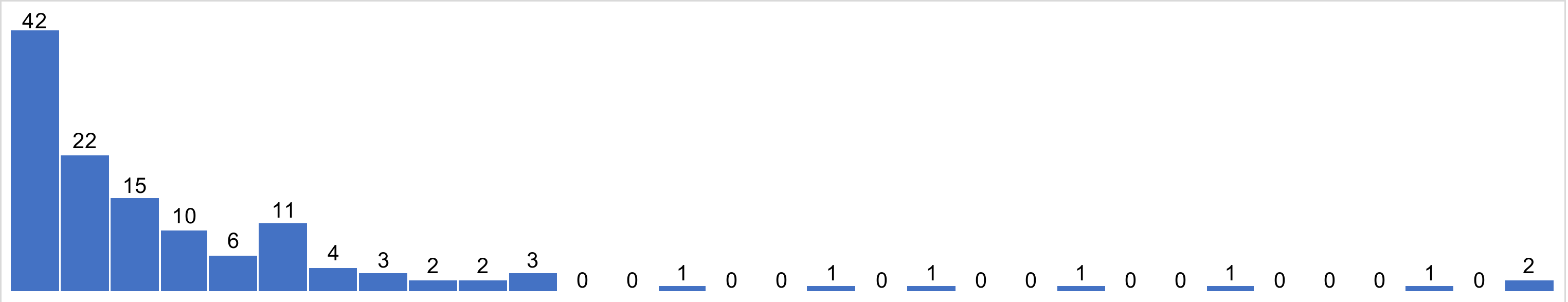}

  \caption{The average lumitexel sent to a particular decoder by our gating module (top 4 rows) and the histogram of the probability distribution of a random lumitexel goes to a particular decoder (bottom). A subset of all average lumitexels are displayed due to limited space. At the bottom graph, the horizontal axis indicates the probability of receiving a random lumitexel from the gating module, which is in the range of [0,0.062], and the vertical axis shows the number of decoders.}
  \label{fig:prob}
\end{figure}
To see what lumitexels each decoder is tuned to, we compute in~\figref{fig:prob} the average lumitexel among all that are sent to a particular decoder by our gating module, over 100K randomly sampled lumitexels. We also compute a histogram of the number of decoders, with respect to the probability of receiving a random lumitexel from the gating module, in the same figure. The current distribution is not balanced, as there is no loss explicitly enforcing this property. We have tested the trick proposed in~\cite{Shazeer:2017:OUTRAGEOUSLY}, but do not find substantial differences in loss. More investigations into this issue will be interesting future work.

\begin{figure*}
    \begin{minipage}{7.1in}
        \begin{minipage}{0.02in}
        \hspace{0.02in}        	
      \end{minipage}	
            \begin{minipage}{7in}
                \centering
                \begin{minipage}{1.33in}
                    \centering
                    {\small Diffuse Albedo}
                \end{minipage}		
                \begin{minipage}{1.33in}
                    \centering
                    {\small Specular Albedo}
                \end{minipage}		
                \begin{minipage}{1.33in}
                    \centering
                    {\small Normal}
                \end{minipage}		
                \begin{minipage}{1.33in}
                    \centering
                    {\small Tangent}
                \end{minipage}		
                \begin{minipage}{1.33in}
                    \centering
                    {\small Roughnesses}
                \end{minipage}		
            \end{minipage}
    \end{minipage}       

        \begin{minipage}{7.1in}
        \begin{minipage}{0.02in}	
            \centering
            \rotatebox{90}{\small \textsc{Fabric}}
        \end{minipage}	
        \begin{minipage}{7in}
            \centering
            \includegraphics[width=1.33in]{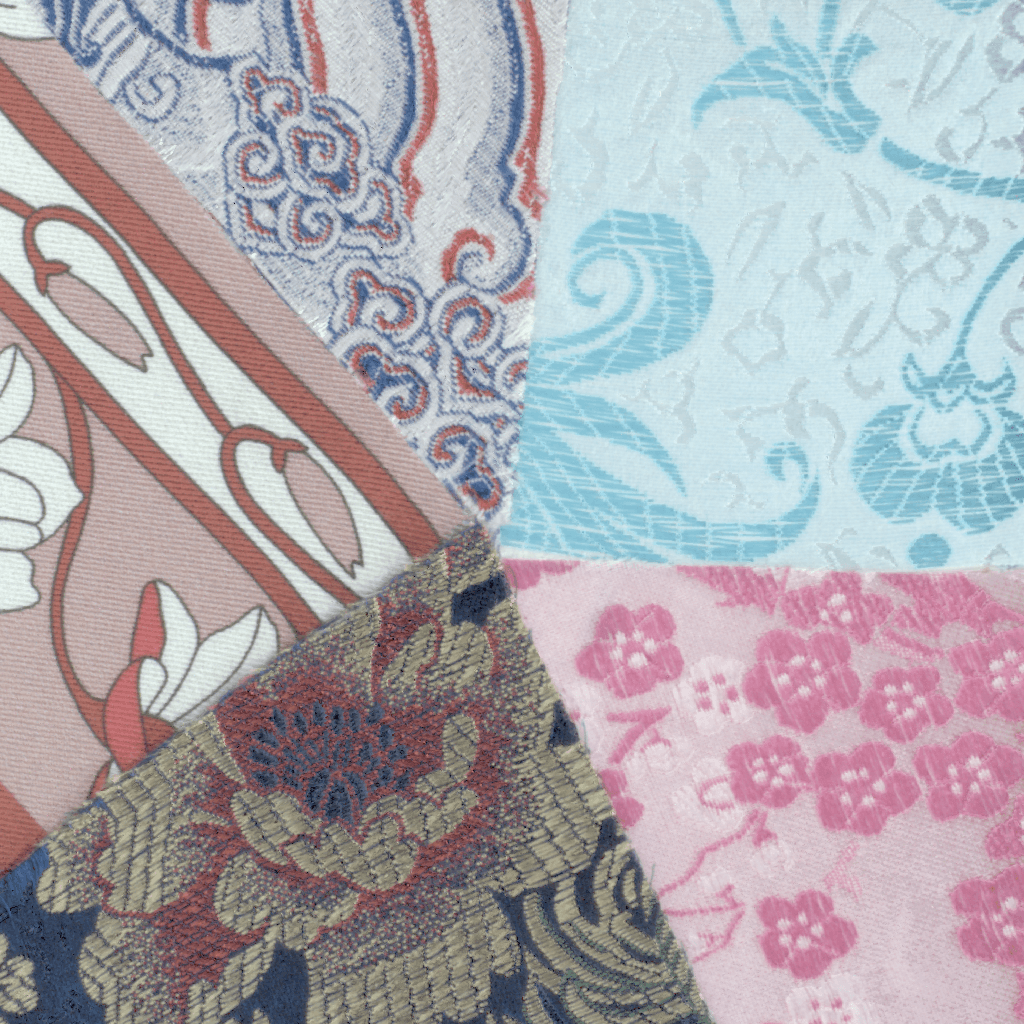}
            \includegraphics[width=1.33in]{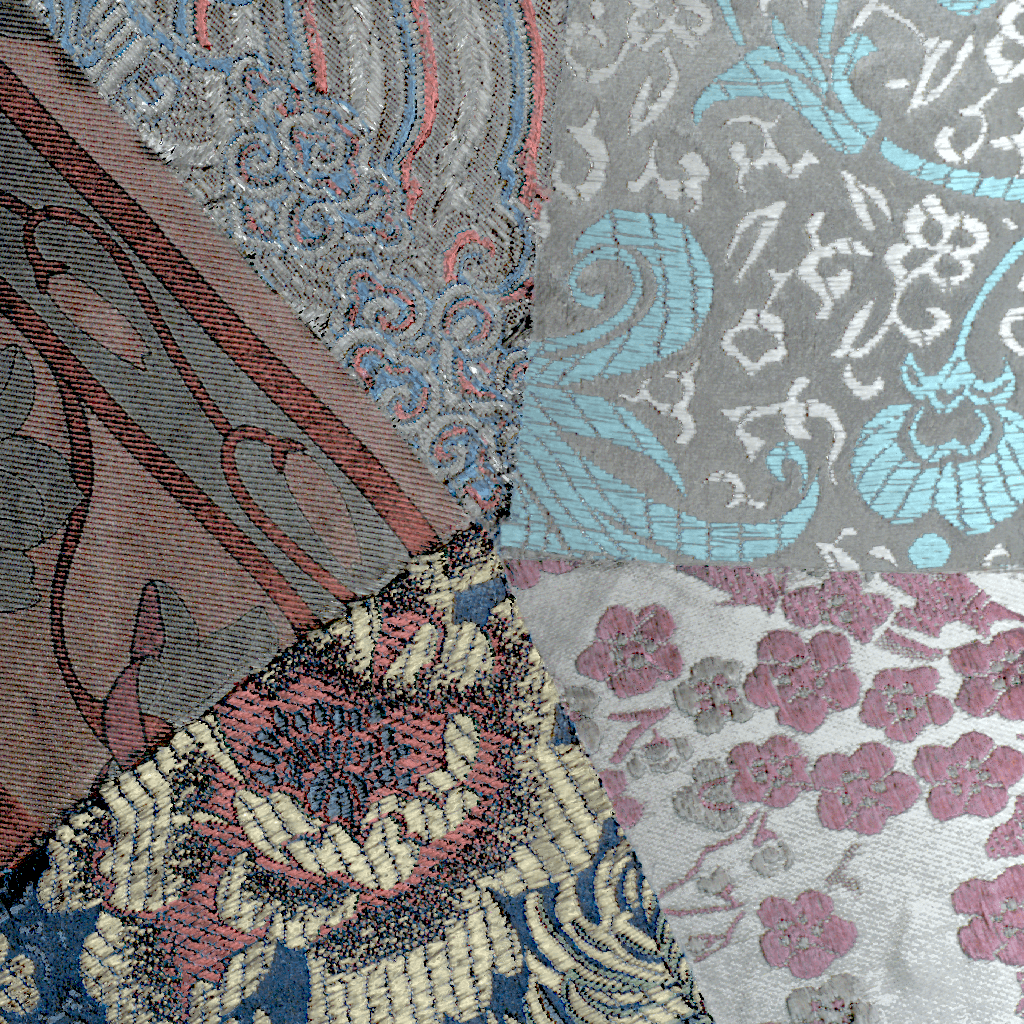}
            \includegraphics[width=1.33in]{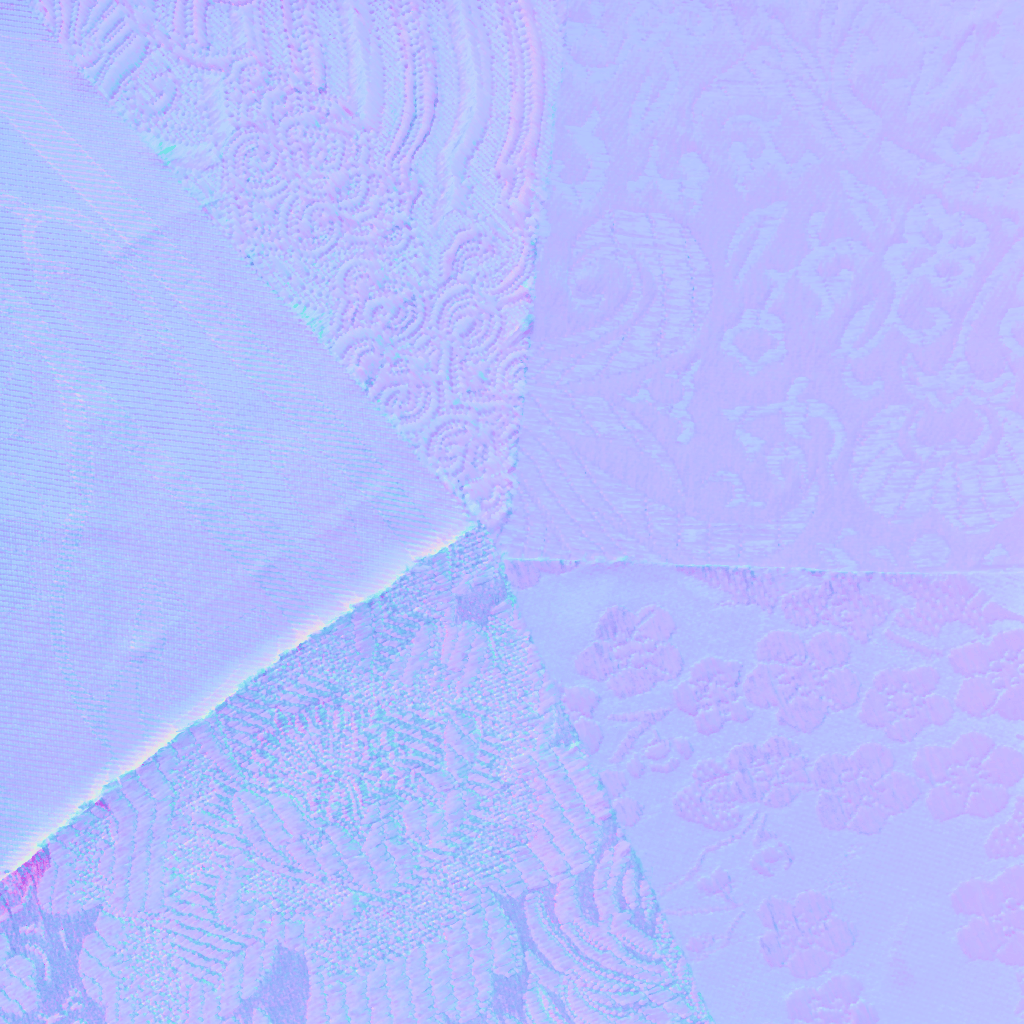}
            \includegraphics[width=1.33in]{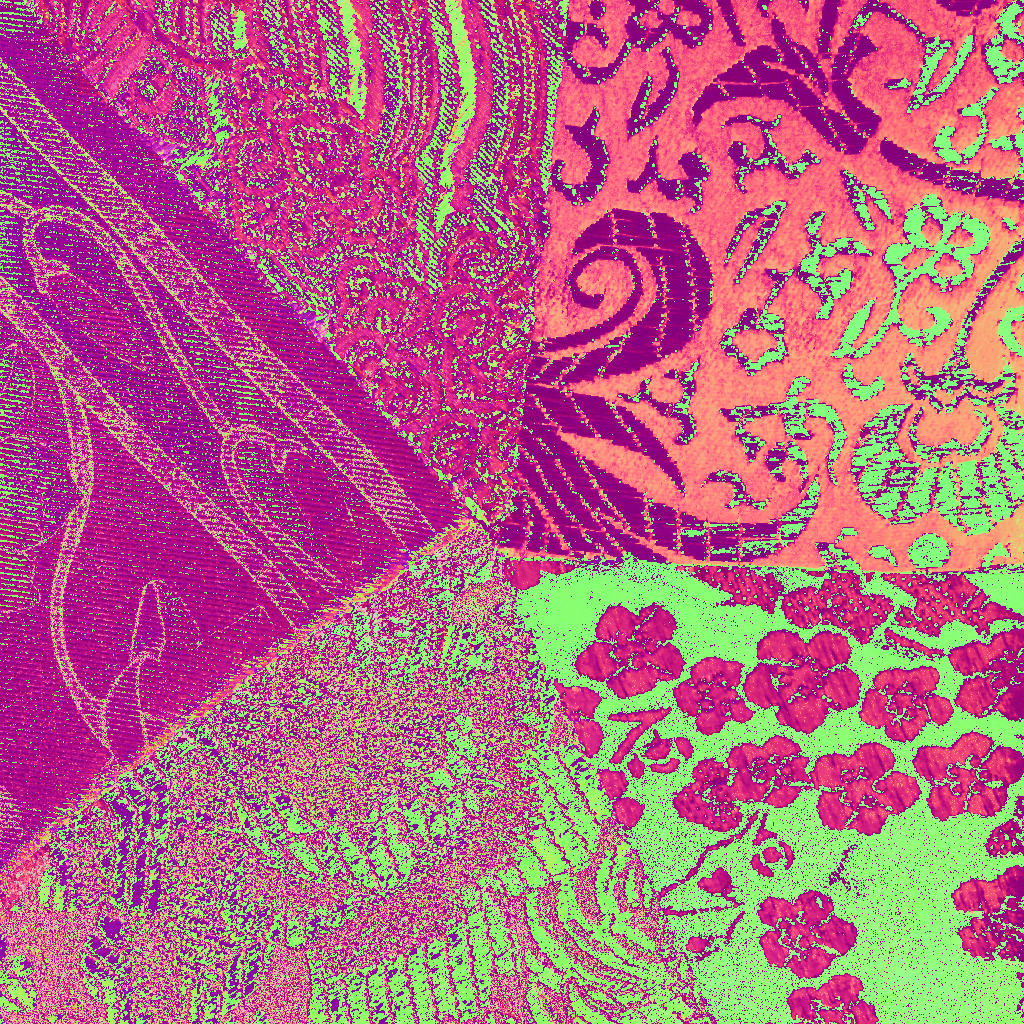}
            \includegraphics[width=1.33in]{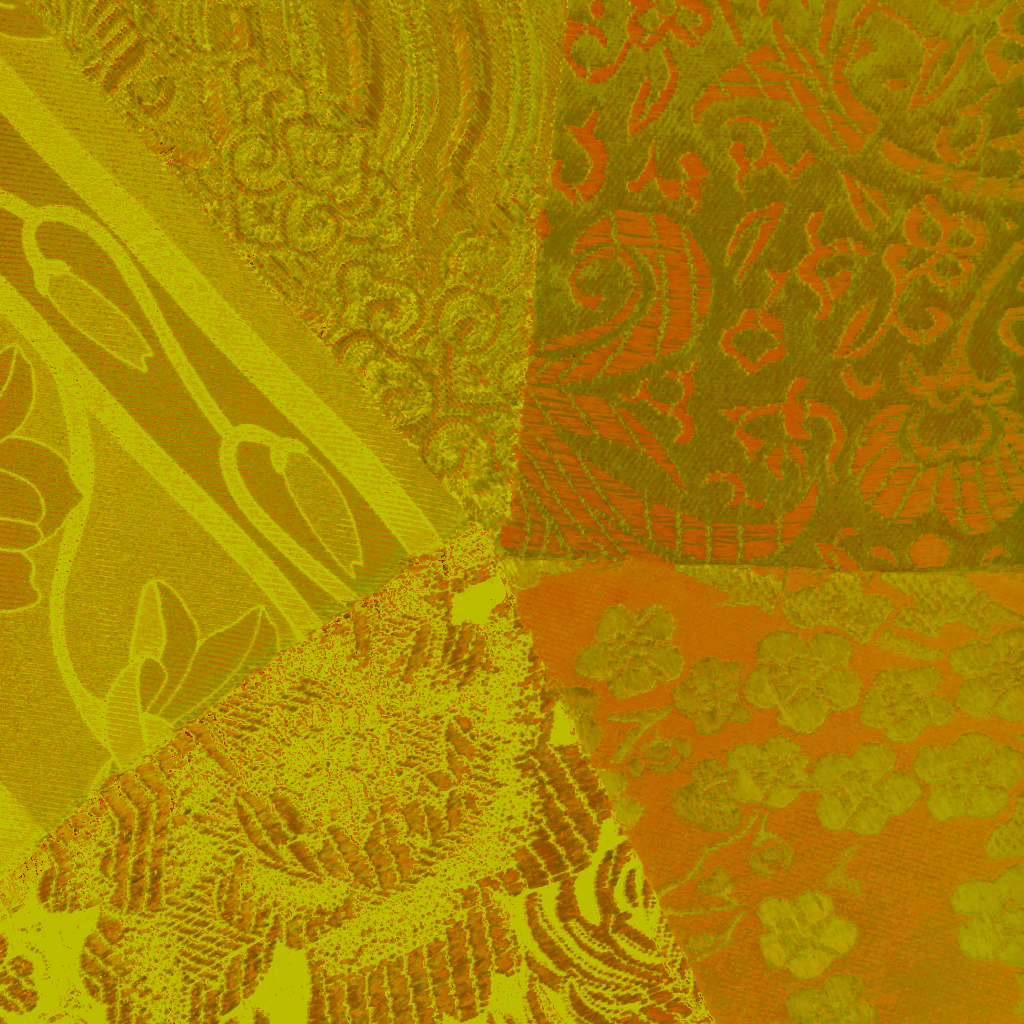}
        \end{minipage}
        \end{minipage}
          
        \begin{minipage}{7.1in}
        \begin{minipage}{0.02in}	
            \centering
            \rotatebox{90}{\small \textsc{Leather}}
        \end{minipage}	
        \begin{minipage}{7in}
            \centering
            \includegraphics[width=1.33in]{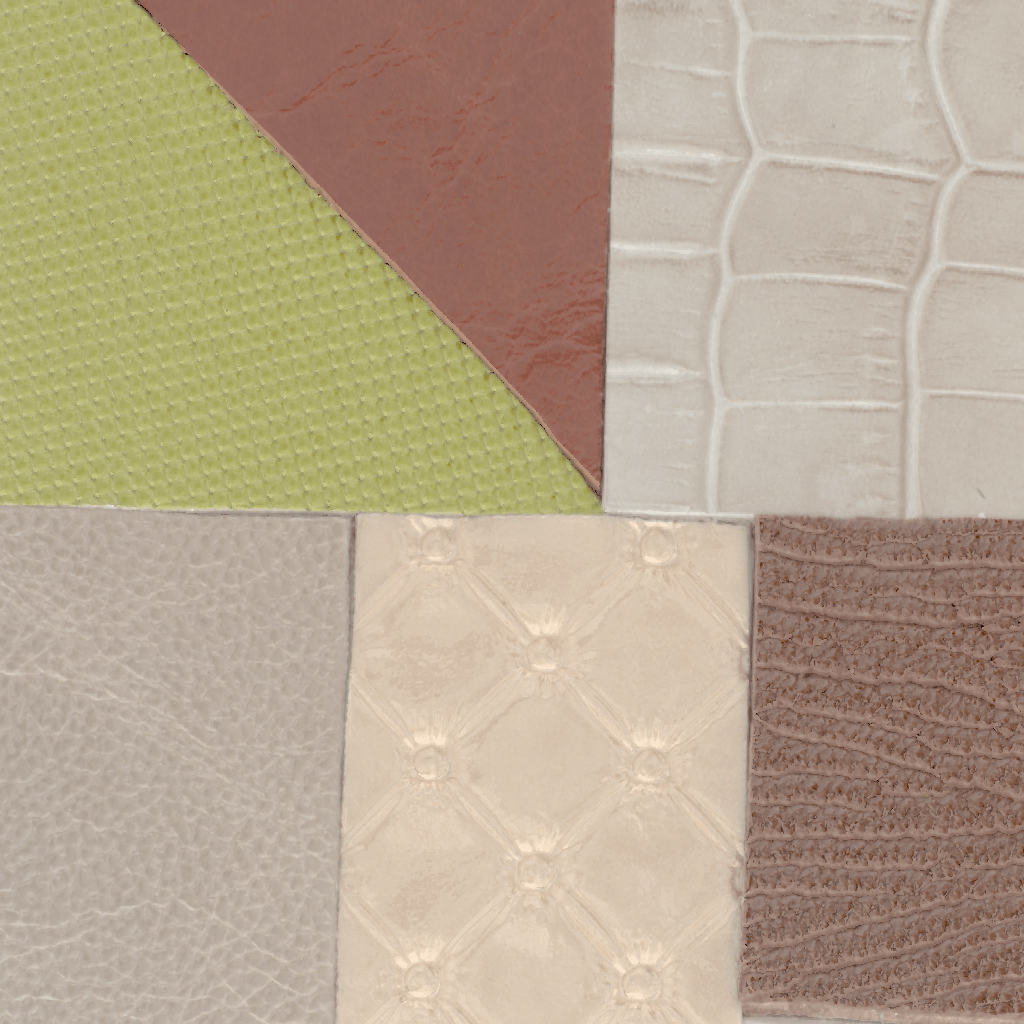}
            \includegraphics[width=1.33in]{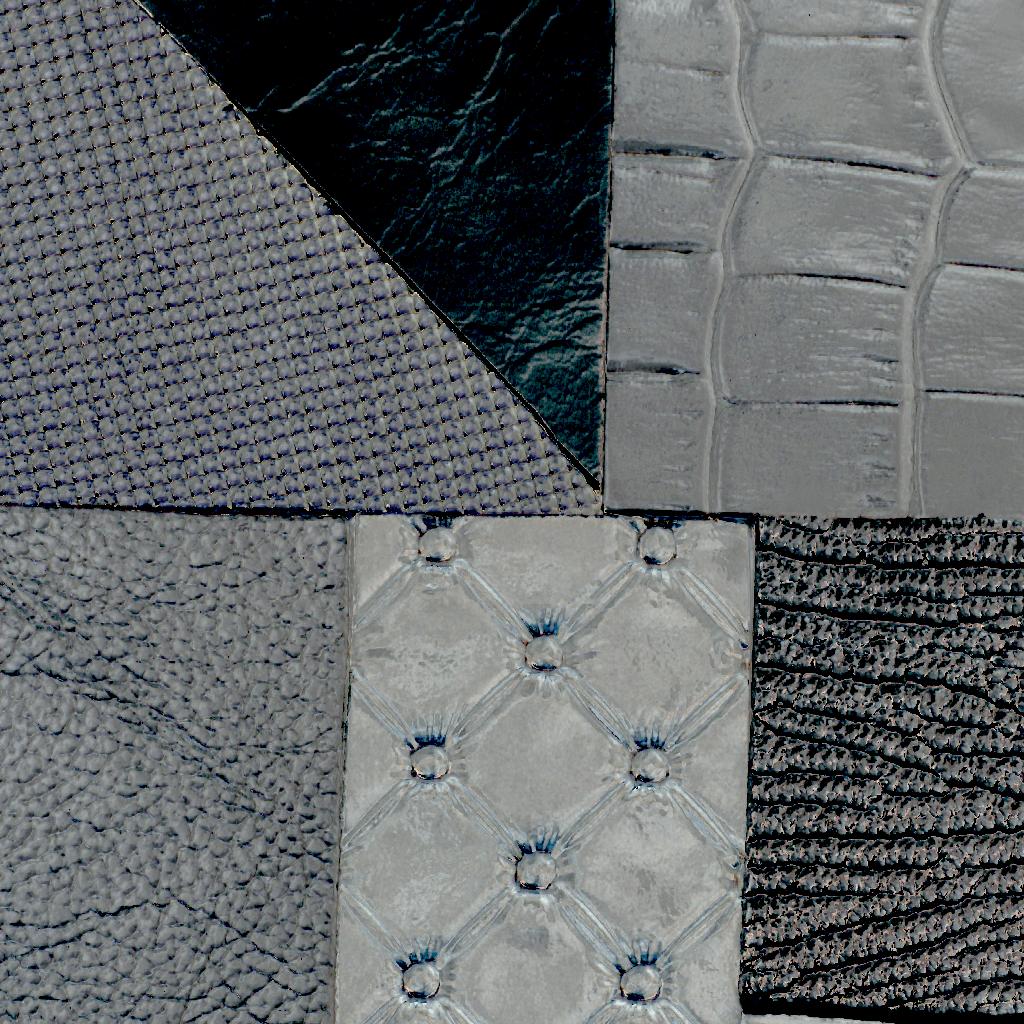}
            \includegraphics[width=1.33in]{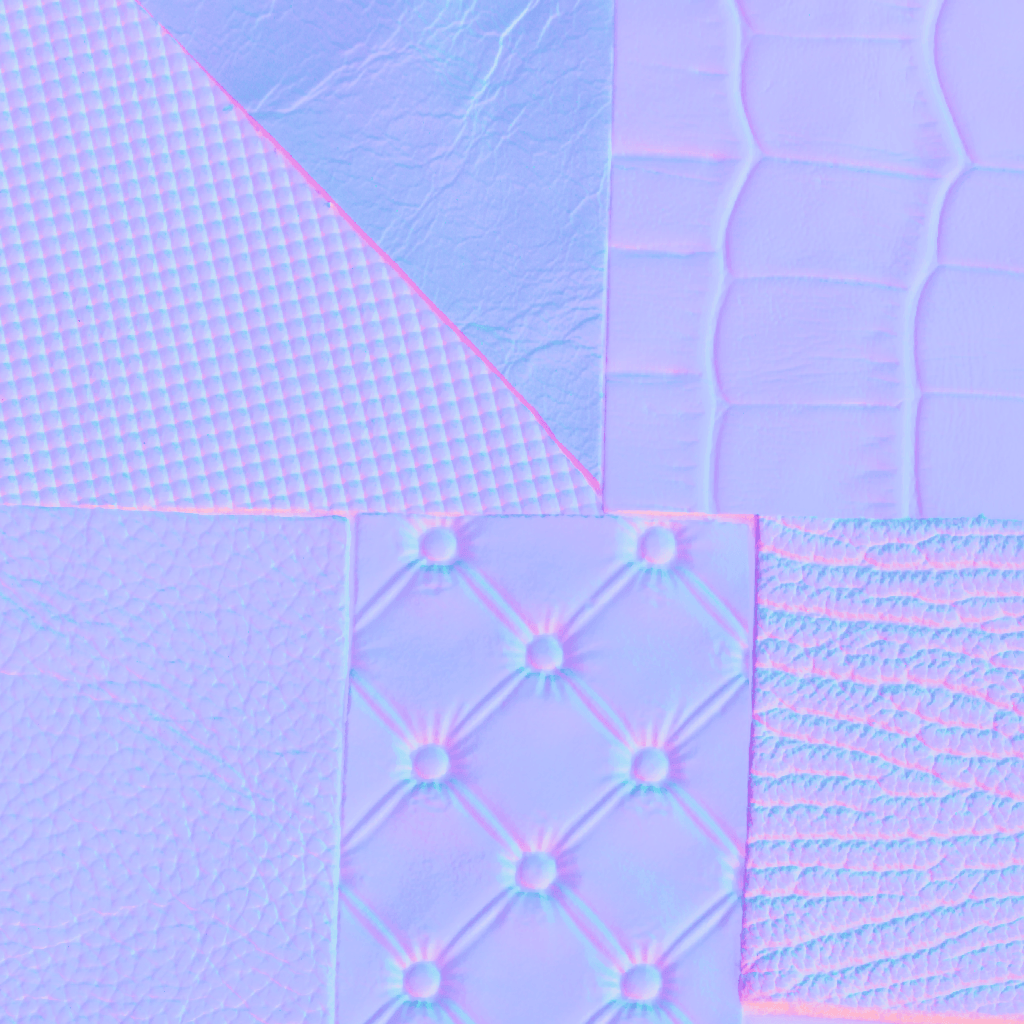}
            \includegraphics[width=1.33in]{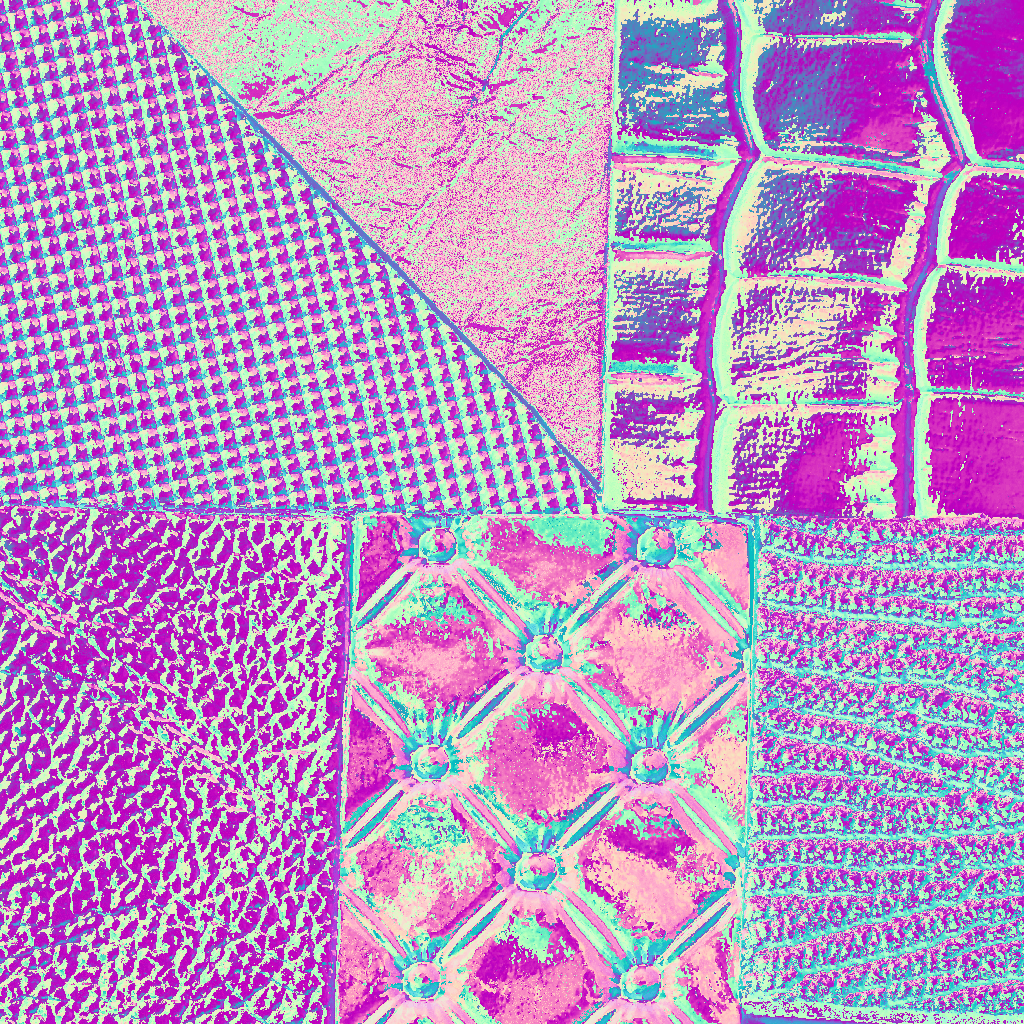}
            \includegraphics[width=1.33in]{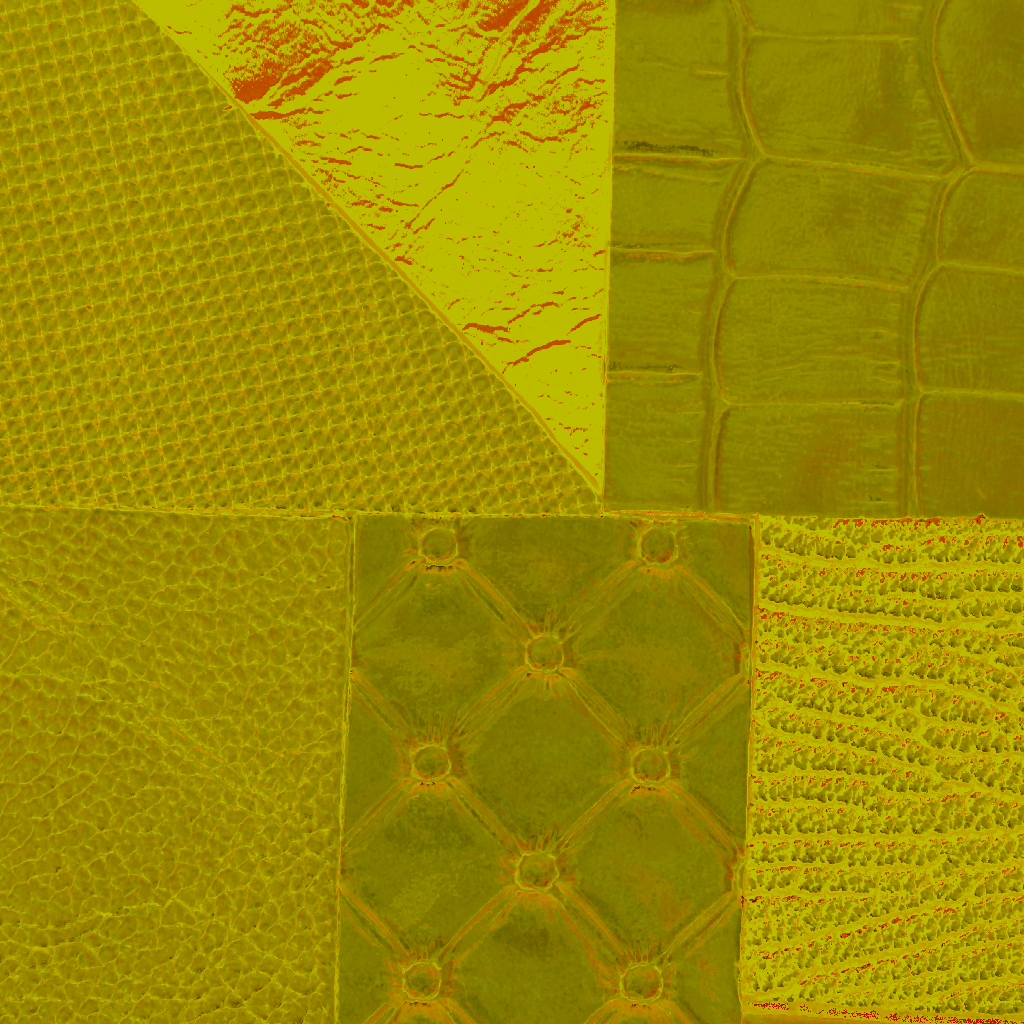}
        \end{minipage}
        \end{minipage}
        
        \begin{minipage}{7.1in}
        \begin{minipage}{0.02in}	
            \centering
            \rotatebox{90}{\small \textsc{Paper}}
        \end{minipage}	
        \begin{minipage}{7in}
            \centering
            \includegraphics[width=1.33in]{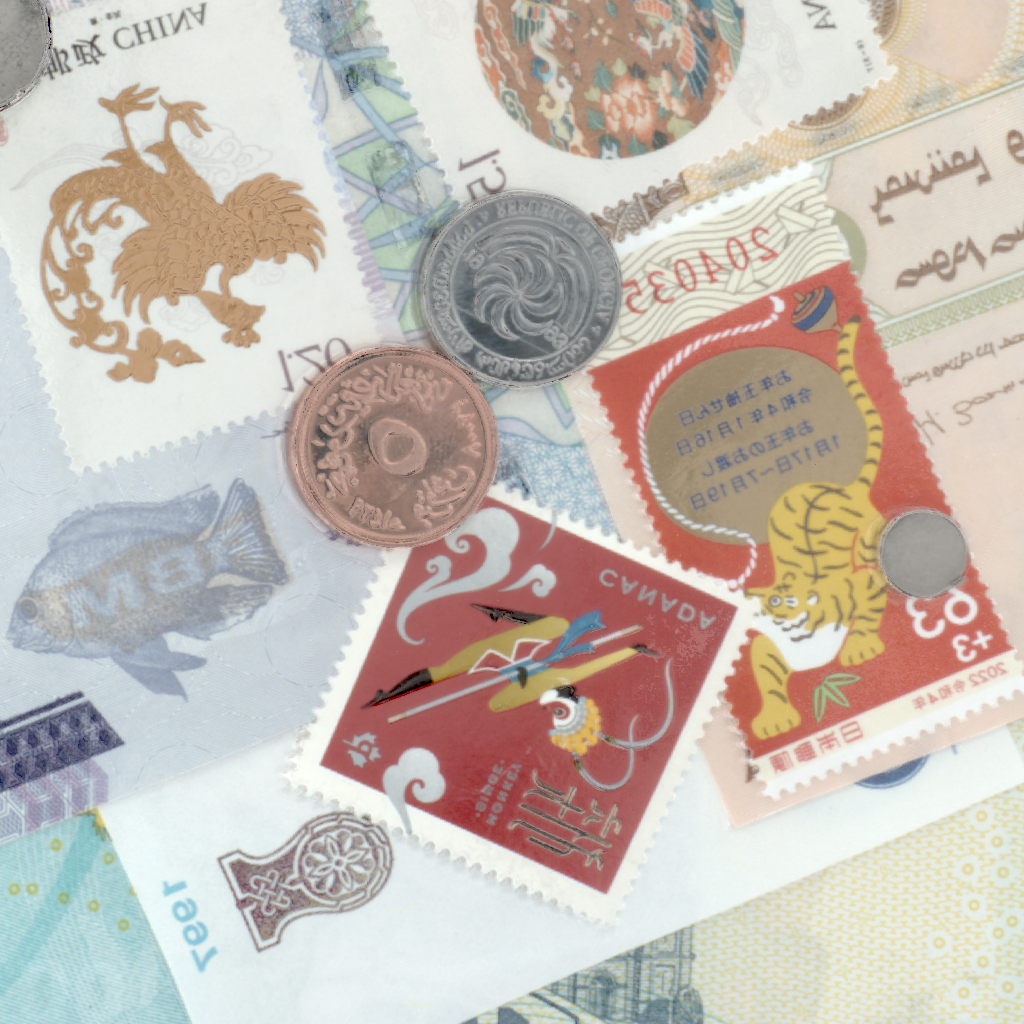}
            \includegraphics[width=1.33in]{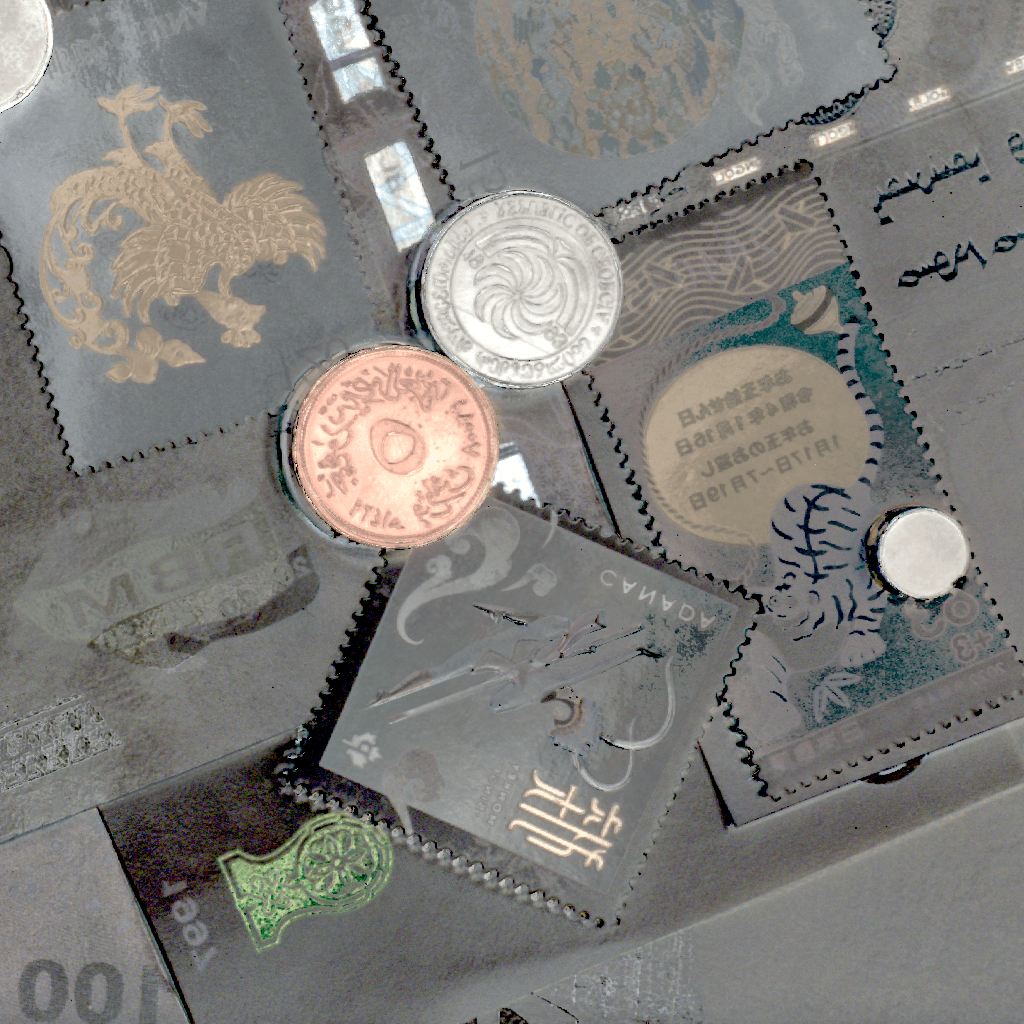}
            \includegraphics[width=1.33in]{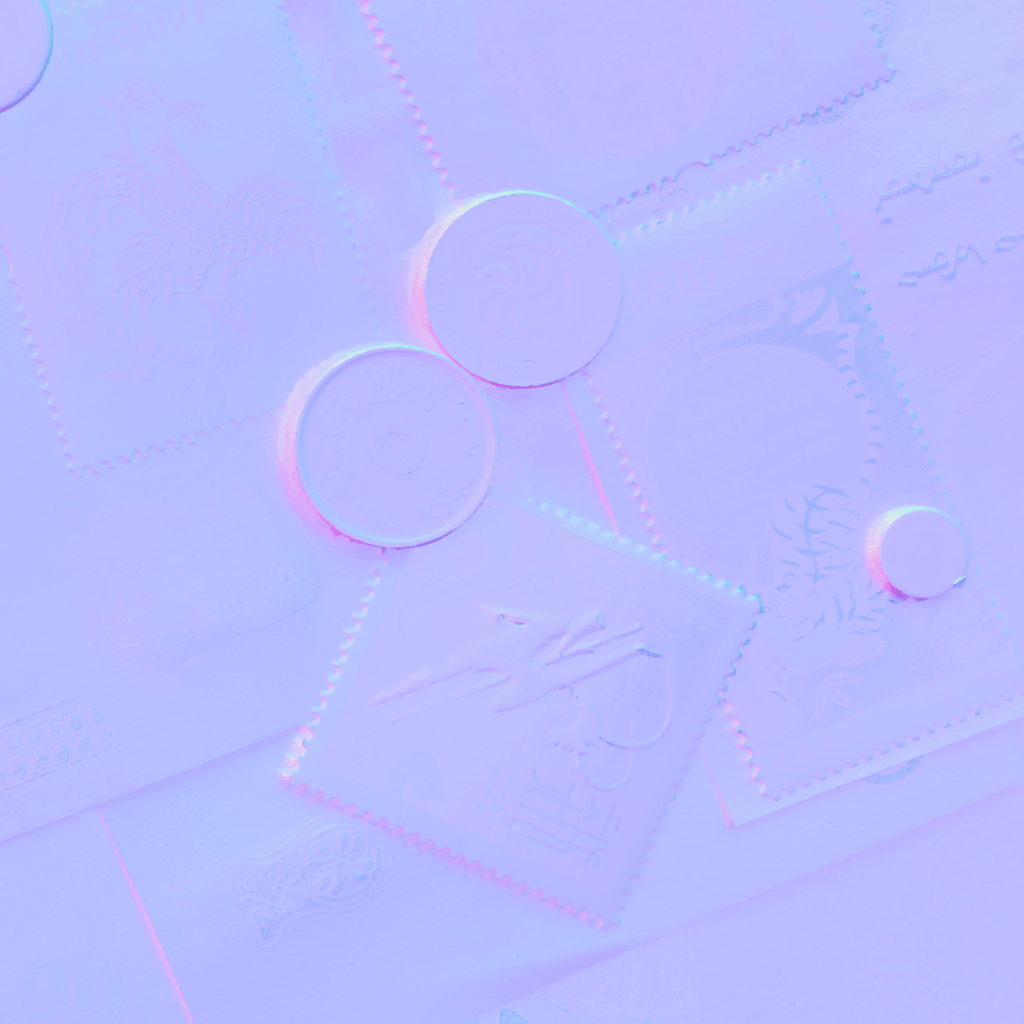}
            \includegraphics[width=1.33in]{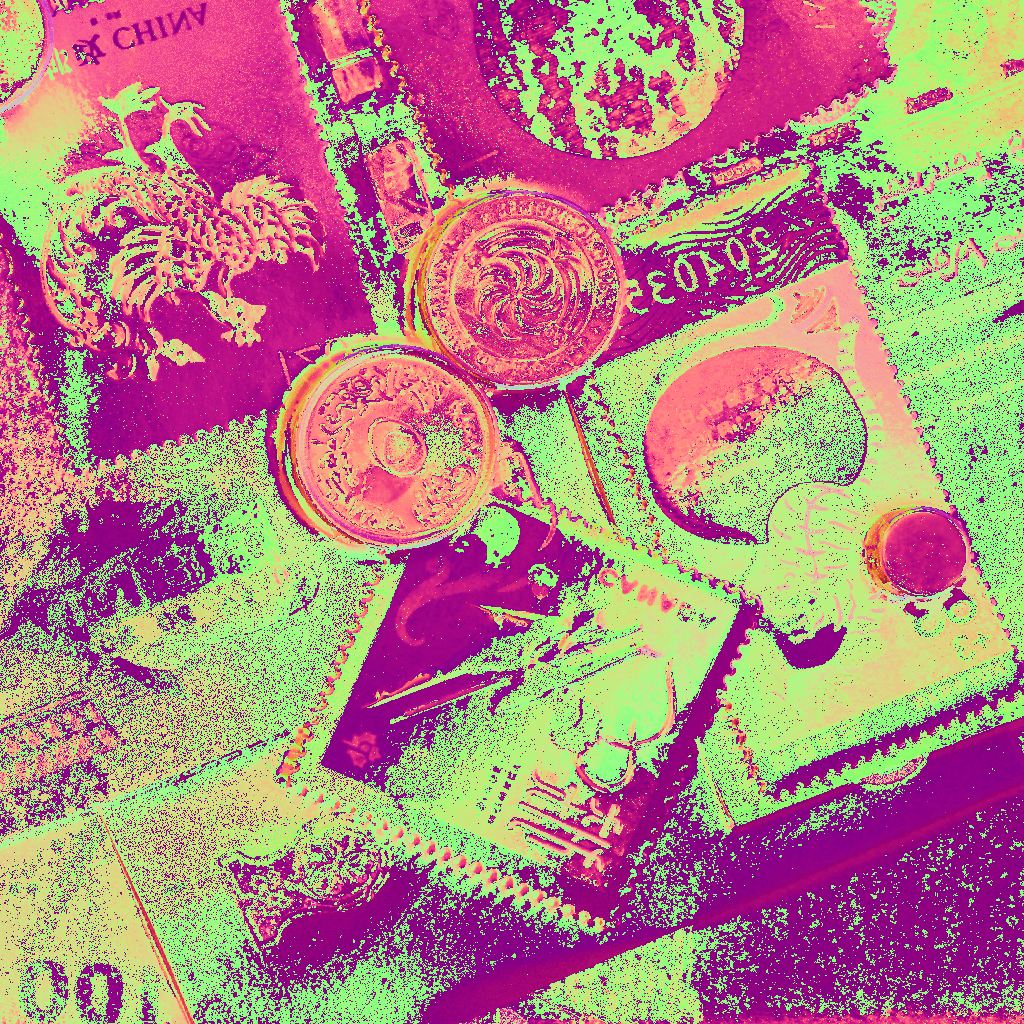}
            \includegraphics[width=1.33in]{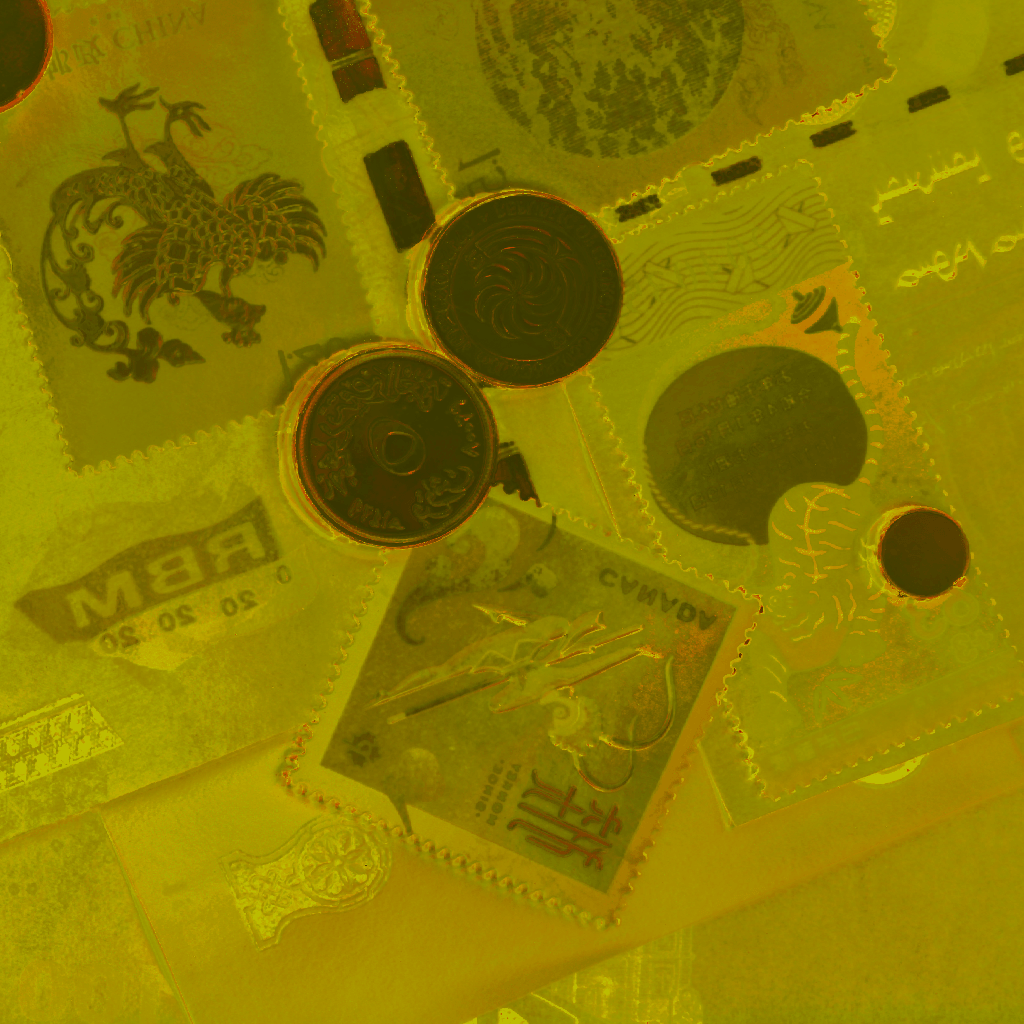}
        \end{minipage}
        \end{minipage}

        \begin{minipage}{7.1in}
        \begin{minipage}{0.02in}	
            \centering
            \rotatebox{90}{\small \textsc{Wood}}
        \end{minipage}	
        \begin{minipage}{7in}
            \centering
            \includegraphics[width=1.33in]{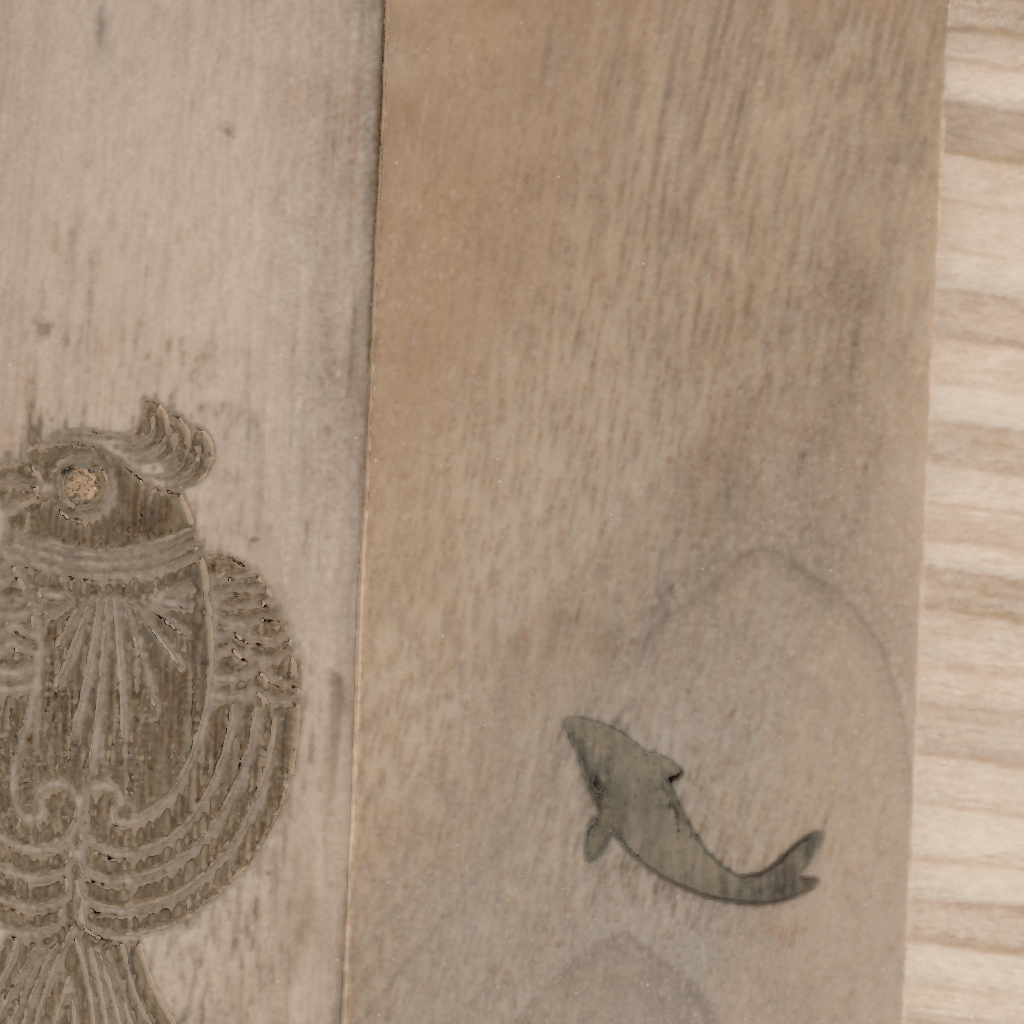}
            \includegraphics[width=1.33in]{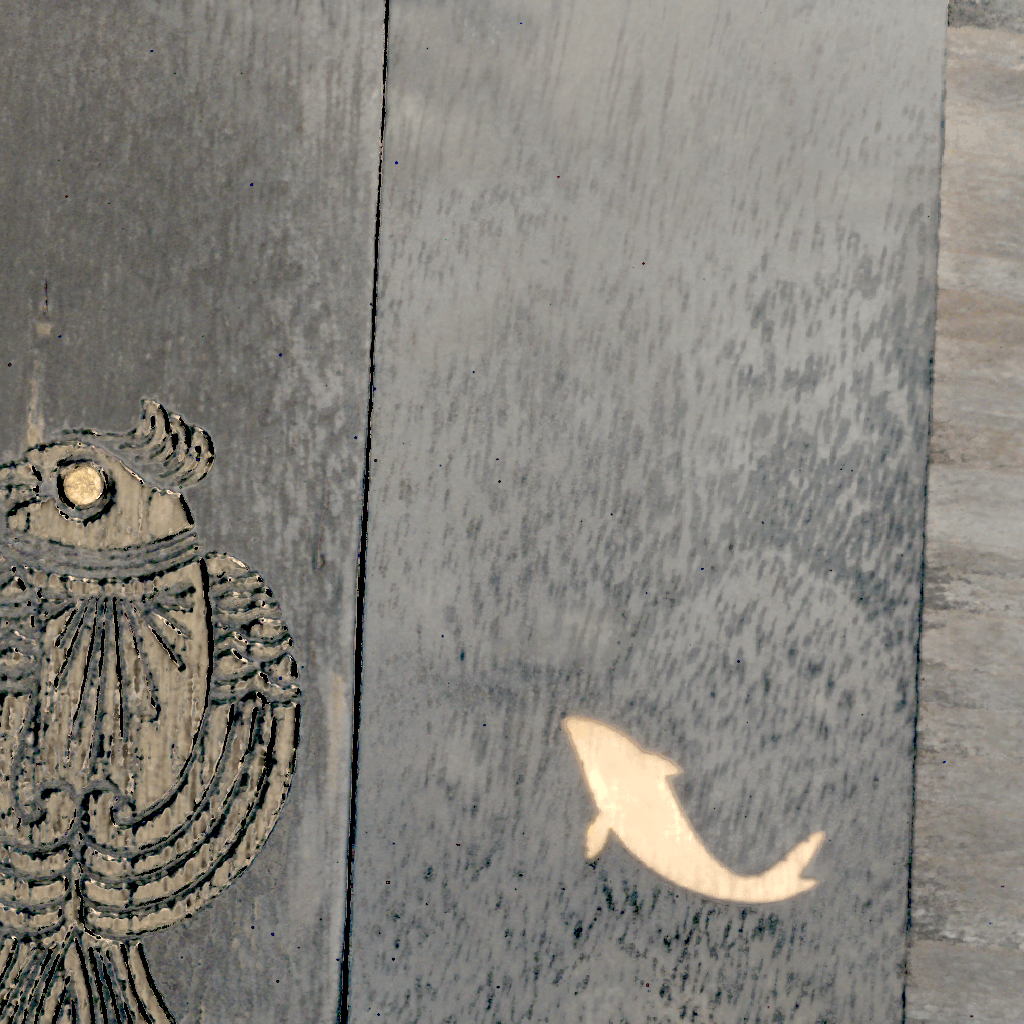}
            \includegraphics[width=1.33in]{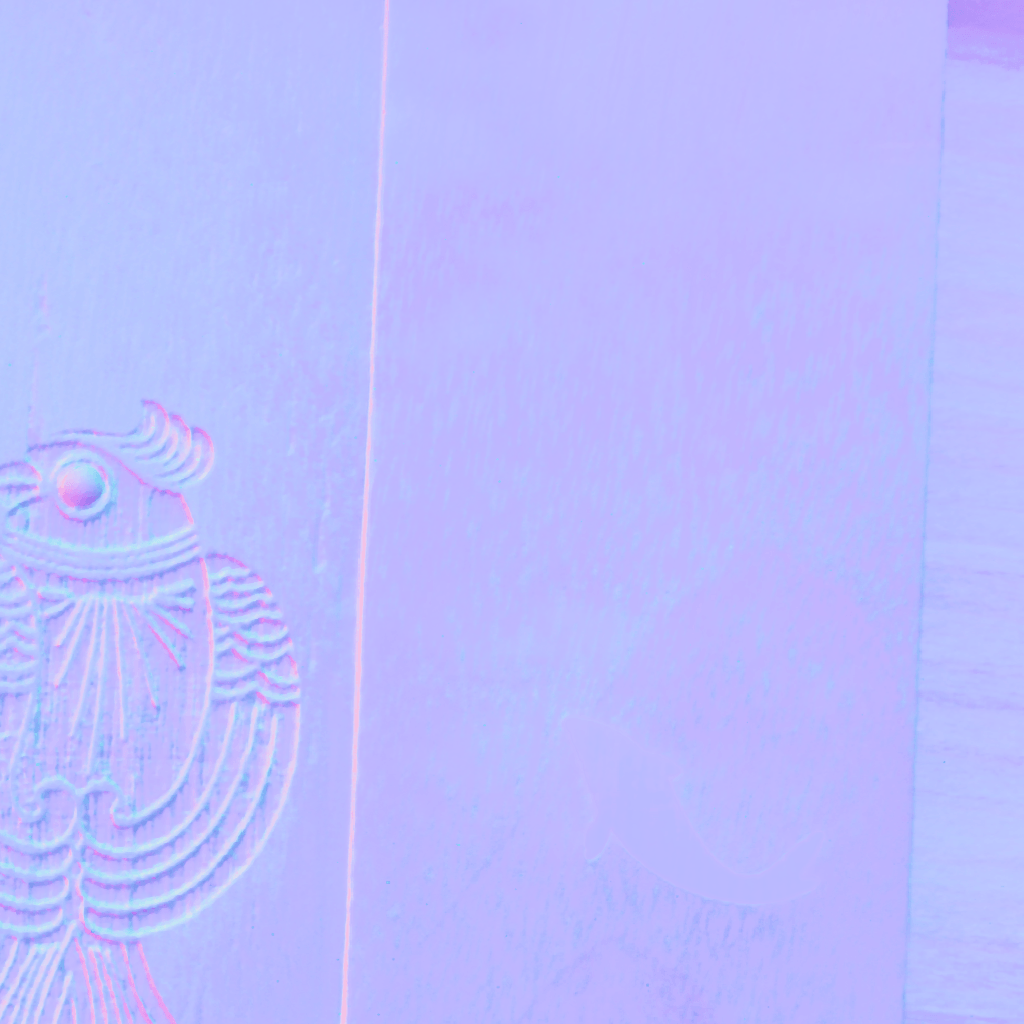}
            \includegraphics[width=1.33in]{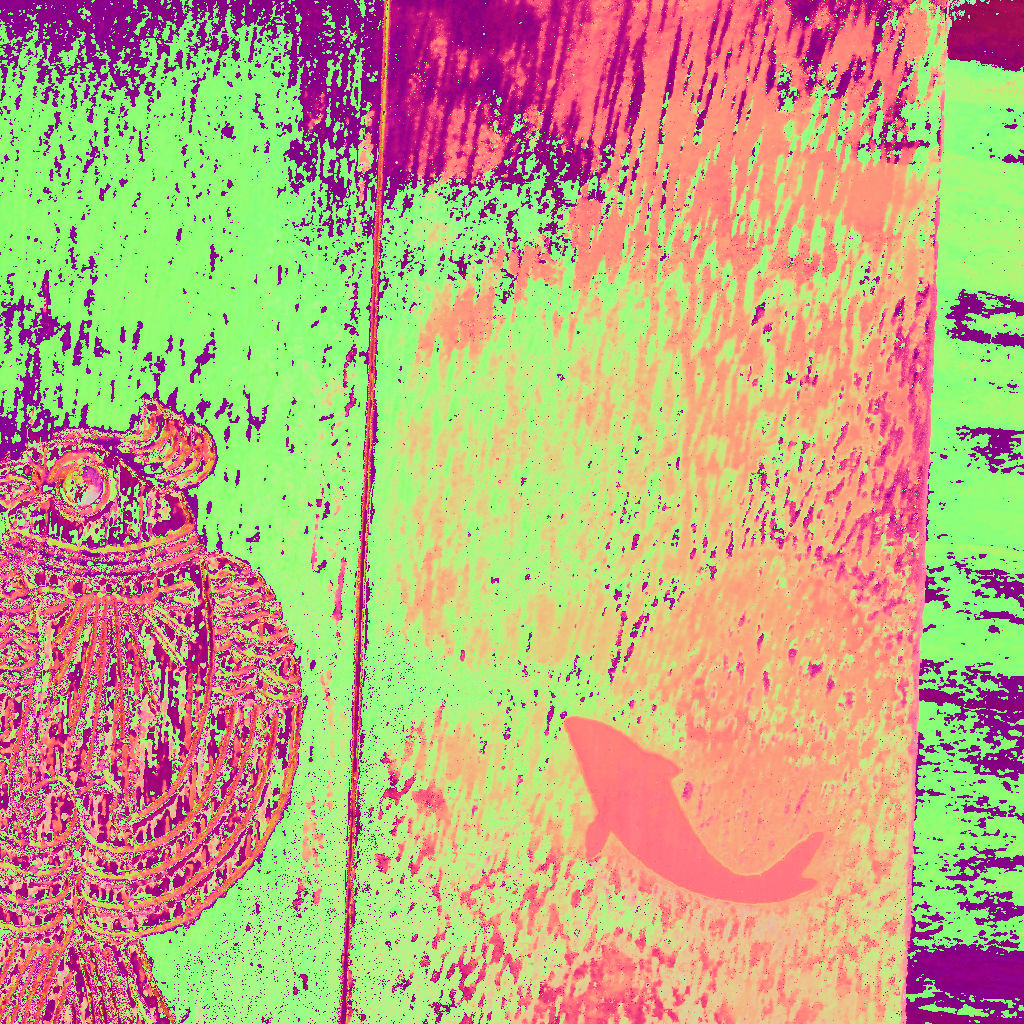}
            \includegraphics[width=1.33in]{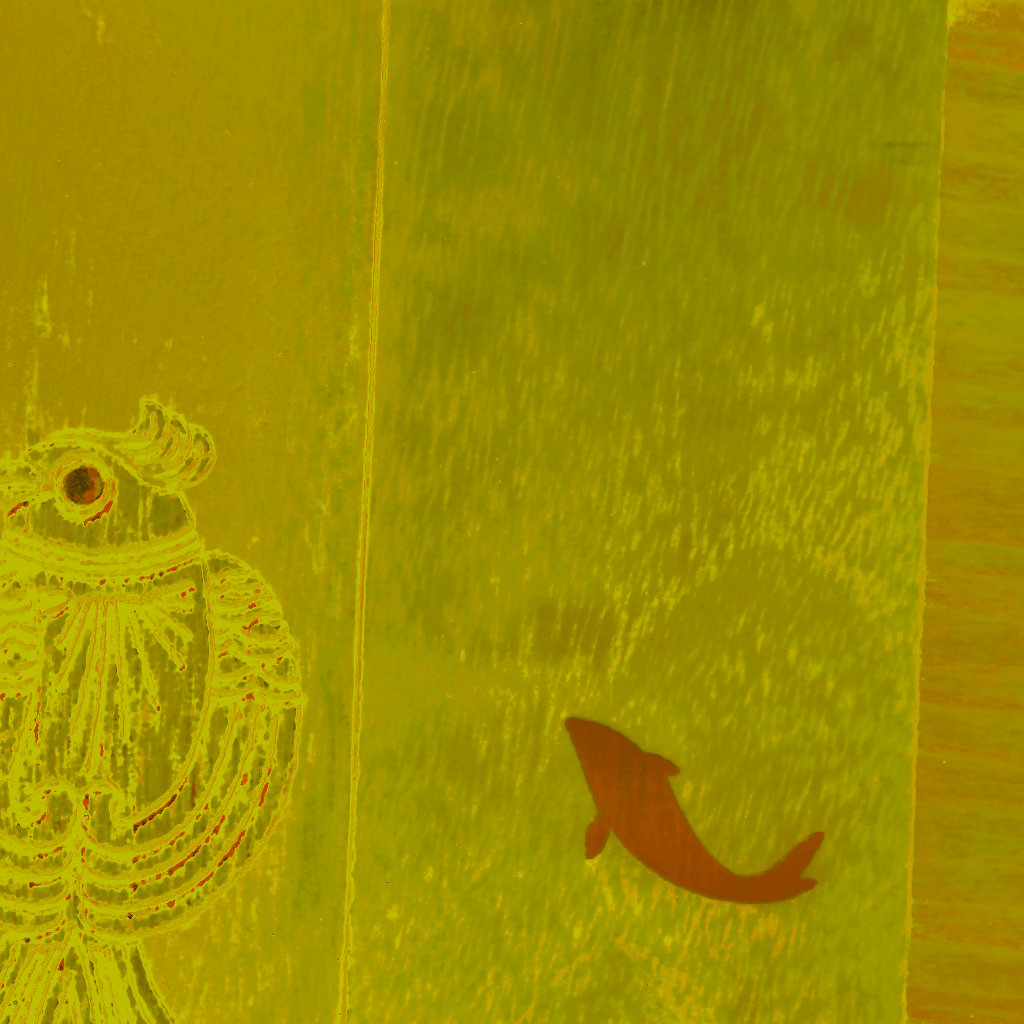}
        \end{minipage}
        \end{minipage}

  \caption{GGX model fitting results from our network (\#=32). Each normal/tangent is added with $(1,1,1)$ and then divided by $2$ to fit to the range of $[0, 1]^3$ for visualization. The roughness $\alpha_{x}/\alpha_{y}$ is visualized in the red/green channel.}
    \label{fig:texture}
\end{figure*}

In~\figref{fig:texture}, we show reflectance fitting results of 4 physical sample sets with our network (\#=32), in the form of texture maps that represent GGX parameters. Our network separates the diffuse and specular reflections, estimates challenging anistropic reflectance and produces high-quality normal maps. It is interesting to observe that how the highly complex appearance on the banknotes in the \textsc{Paper} set is modeled by our approach. In addition, please refer to the accompanying video for rendering results of the sample sets with novel view and lighting conditions.

\subsection{Comparisons}
We validate our results against photographs, and compare with LDAE~\cite{Kang:2018:AUTO} with the \textbf{same} number of lighting patterns(\#=32) in~\figref{fig:validation}. In all cases, our network produces results that more closely resemble the corresponding photographs with a novel lighting condition not used in training, compared with LDAE; superior quantitative errors in SSIM are also reported, demonstrating our improved efficiency.

\begin{figure}
    \centering
    \begin{minipage}{0.32\linewidth}
        \centering
        \small Photo
    \end{minipage}		
    \begin{minipage}{0.32\linewidth}
        \centering
        \small Ours(\#=12)
    \end{minipage}		
    \begin{minipage}{0.32\linewidth}
        \centering
        \small LDAE(\#=32)
    \end{minipage}		
    \begin{minipage}{0.32\linewidth}
        \includegraphics[width=\linewidth]{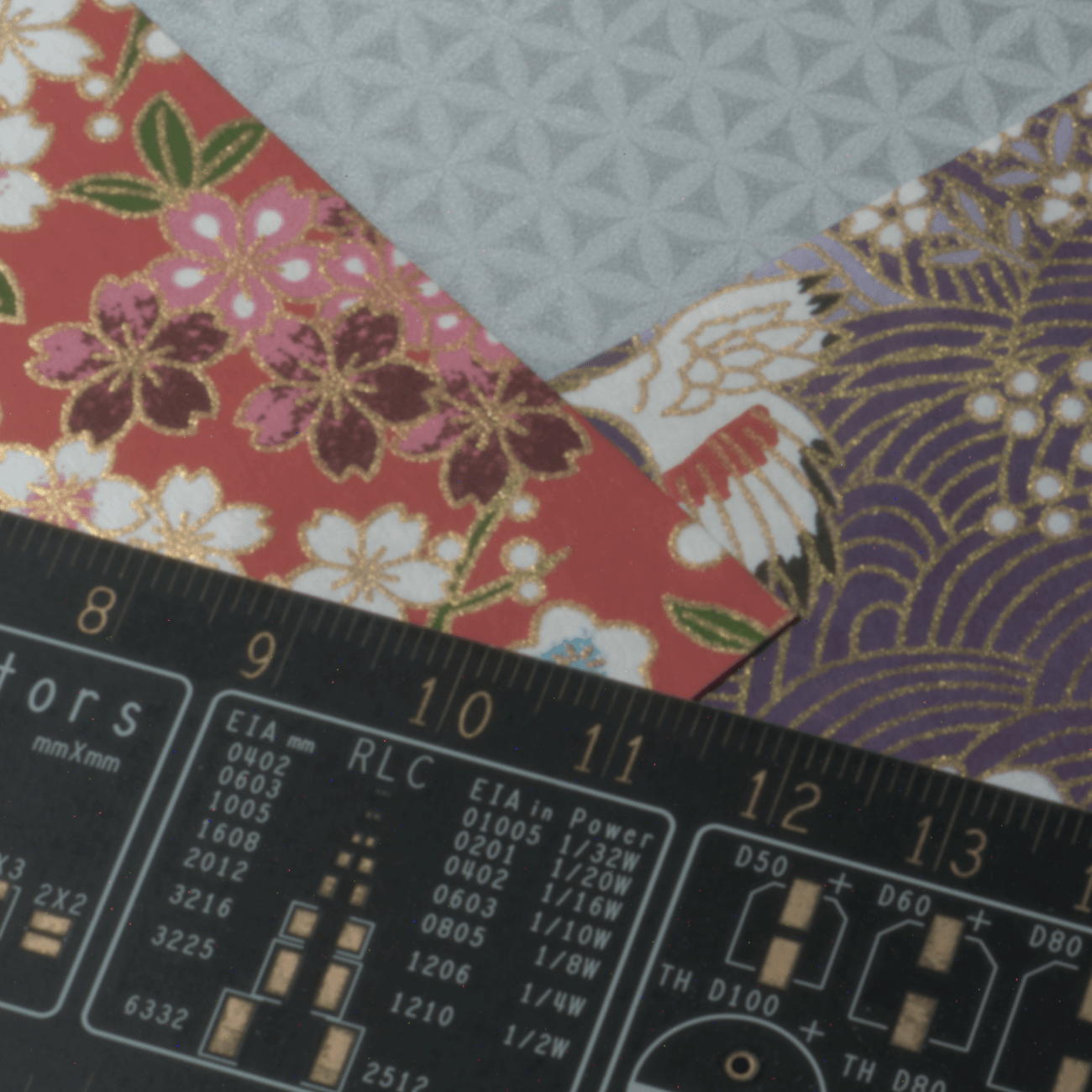}
    \end{minipage}		
    \begin{minipage}{0.32\linewidth}
        \includegraphics[width=\linewidth]{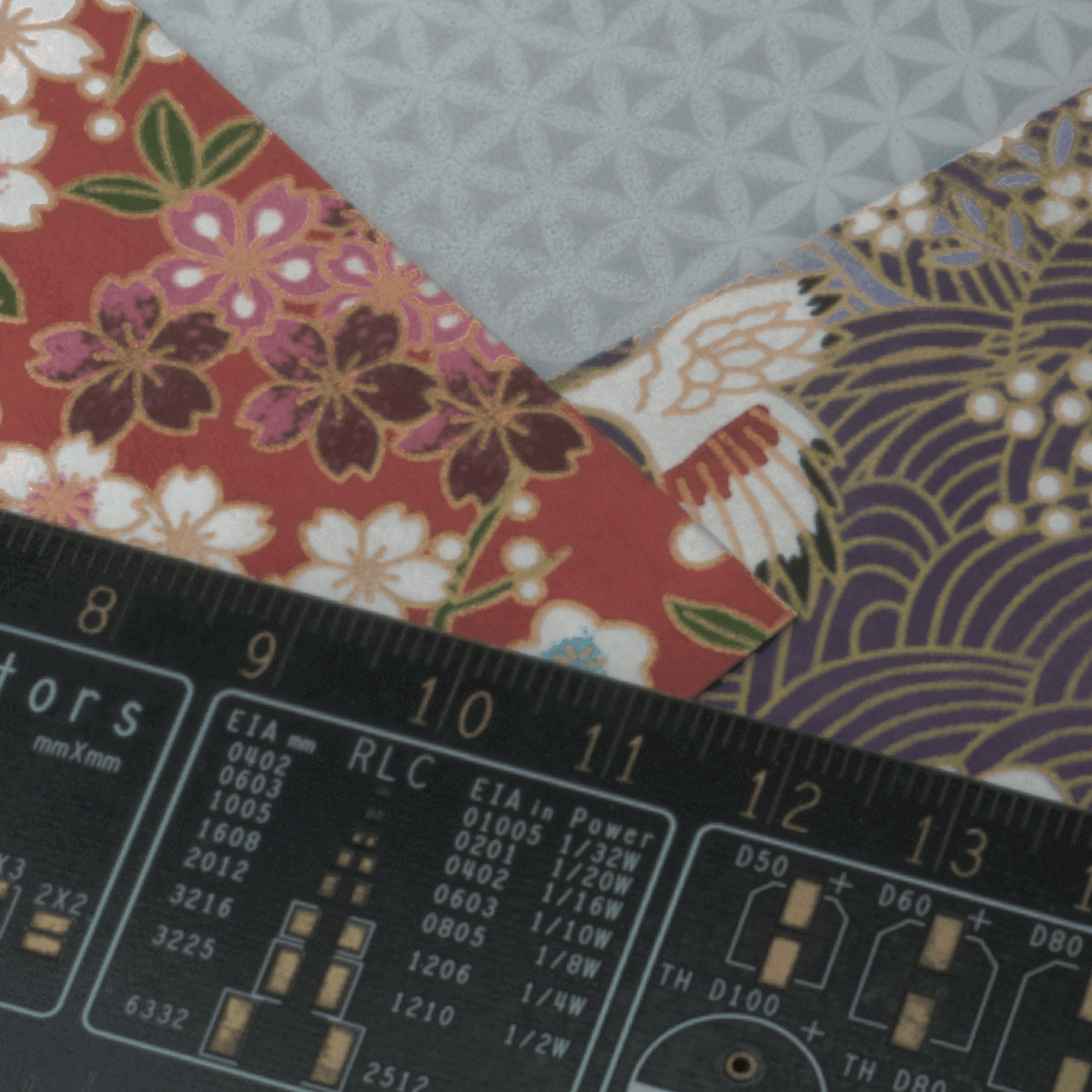}
        \put(-57,5) {\small \color{white} SSIM = 0.94}
    \end{minipage}		
    \begin{minipage}{0.32\linewidth}
        \includegraphics[width=\linewidth]{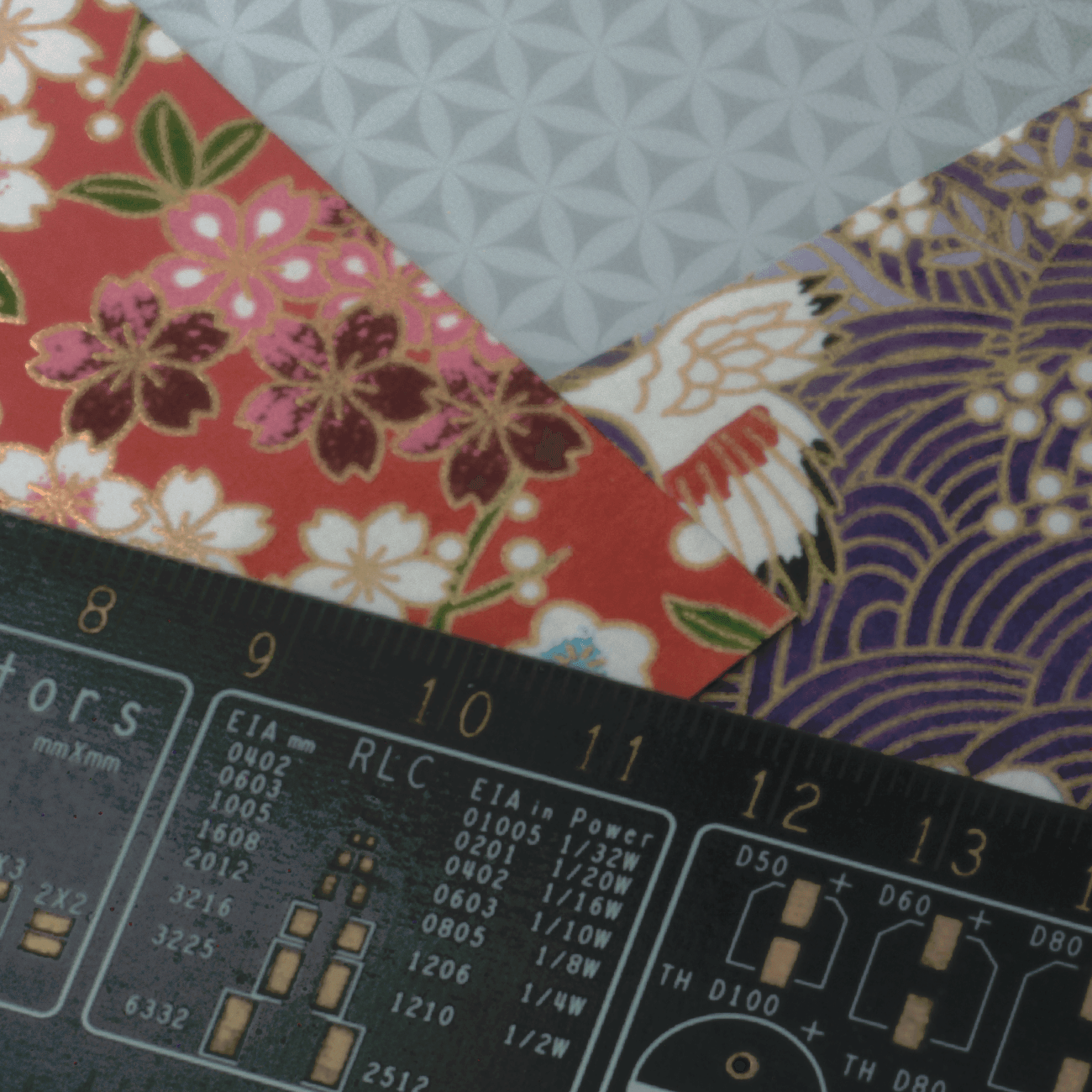}
        \put(-57,5) {\small \color{white} SSIM = 0.93}
    \end{minipage}		
   \caption{Comparison between our network(\#=12) and LDAE(\#=32)~\cite{Kang:2018:AUTO} with similar validation losses. From the left column to right, a photograph of the physical sample set, our result and the result of LDAE.}
   \label{fig:equal_loss}
\end{figure}

In~\figref{fig:equal_loss}, our network($\#$=12) is compared with LDAE($\#$=32), both of which have similar validation losses, according to~\figref{fig:loss}. Our results are comparable to LDAE both qualitatively and quantitatively, with respect to the corresponding photograph. Note that we need only about 1/3 the number of input images, showing a considerable increase in efficiency.

\subsection{Evaluations}
\begin{figure}
     \centering
 	\includegraphics[width = \columnwidth]{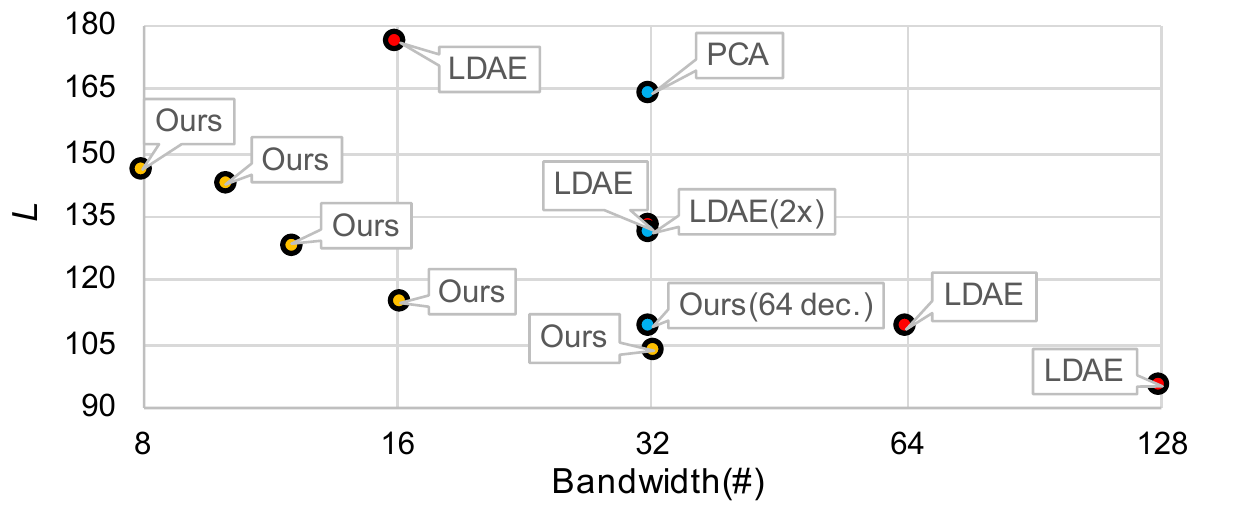}
     	\caption{Comparisons of average prediction qualities of different networks with different parameters. The loss L is computed on the validation dataset. Our networks/LDAEs with different input bandwidths are marked as yellow/red dots, respectively. We also show the losses of several variants in blue dots. Please refer to the main text for details.}
   \label{fig:loss}
\end{figure}

\begin{figure}
    \begin{minipage}{\columnwidth}
        \begin{minipage}{0.32\linewidth}	
            \centering
            {\small Photo}
        \end{minipage}	
        \begin{minipage}{0.32\linewidth}	
            \centering
            {\small Ours(\#=32)}
        \end{minipage}	
        \begin{minipage}{0.32\linewidth}	
            \centering
            {\small LDAE(\#=32)}
        \end{minipage}

        \begin{minipage}{0.32\linewidth}	
            \includegraphics[width=\linewidth]{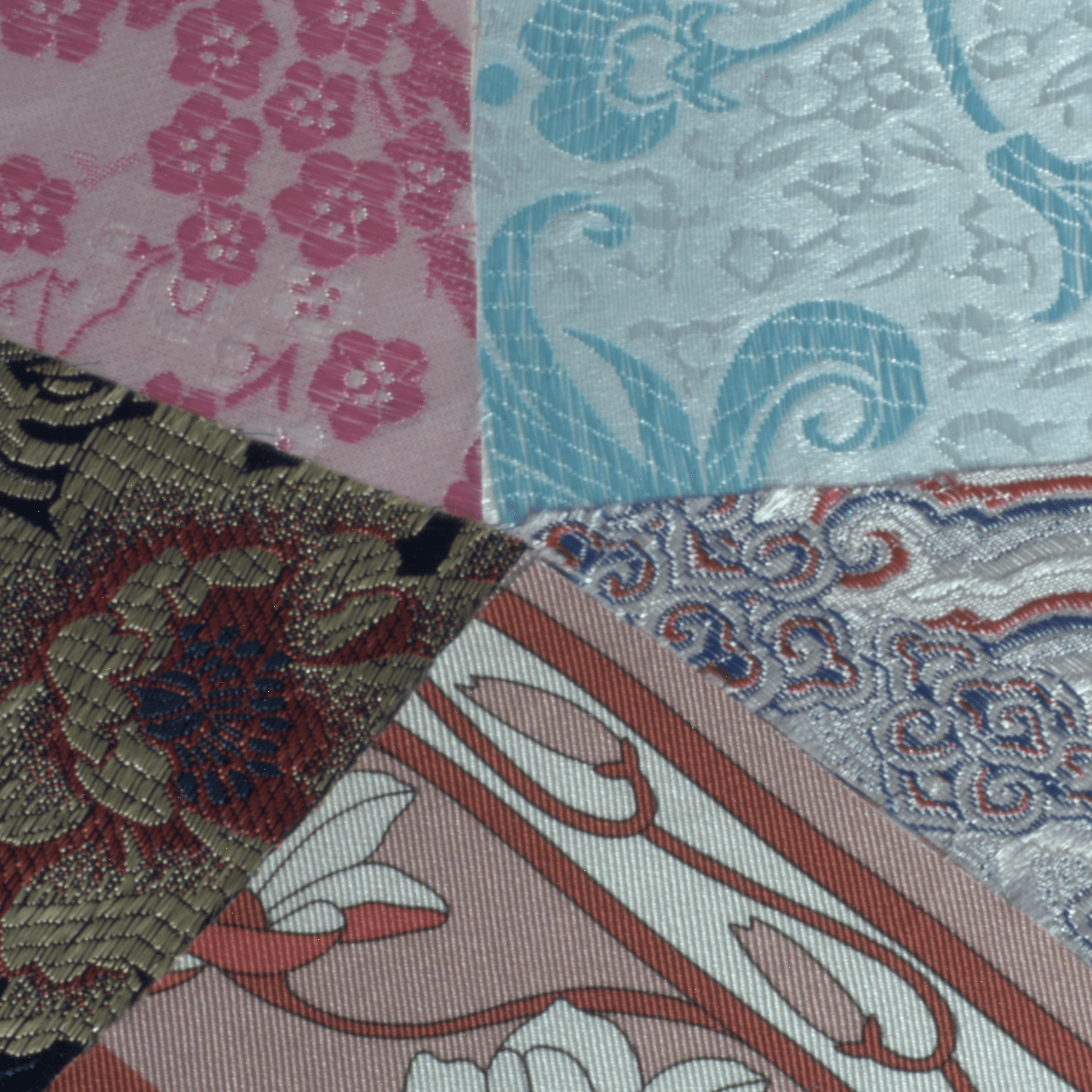}
        \end{minipage}	
        \begin{minipage}{0.32\linewidth}	
            \includegraphics[width=\linewidth]{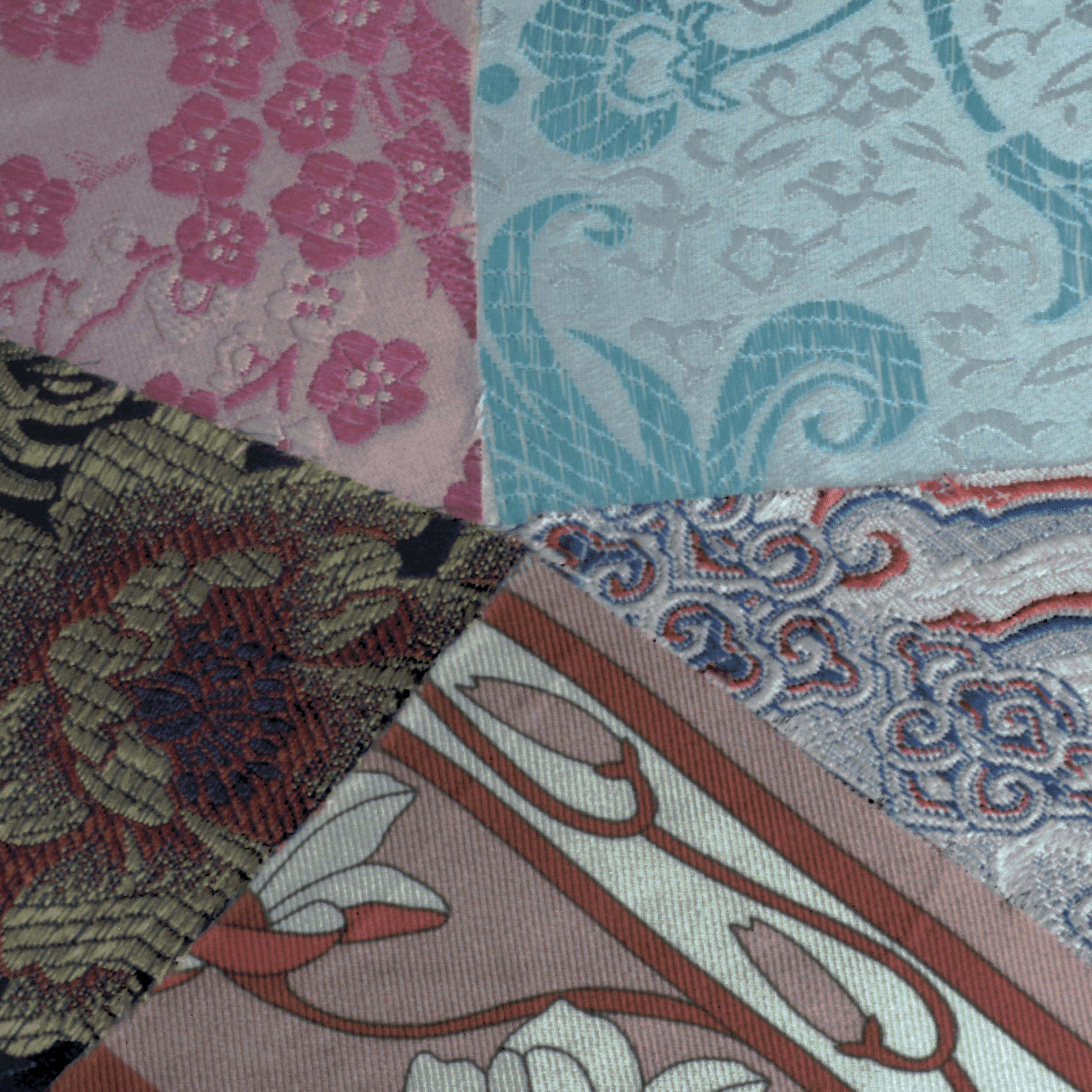}
            \put(-57,3) {\small \color{white} SSIM = 0.90}
        \end{minipage}	
        \begin{minipage}{0.32\linewidth}	
            \includegraphics[width=\linewidth]{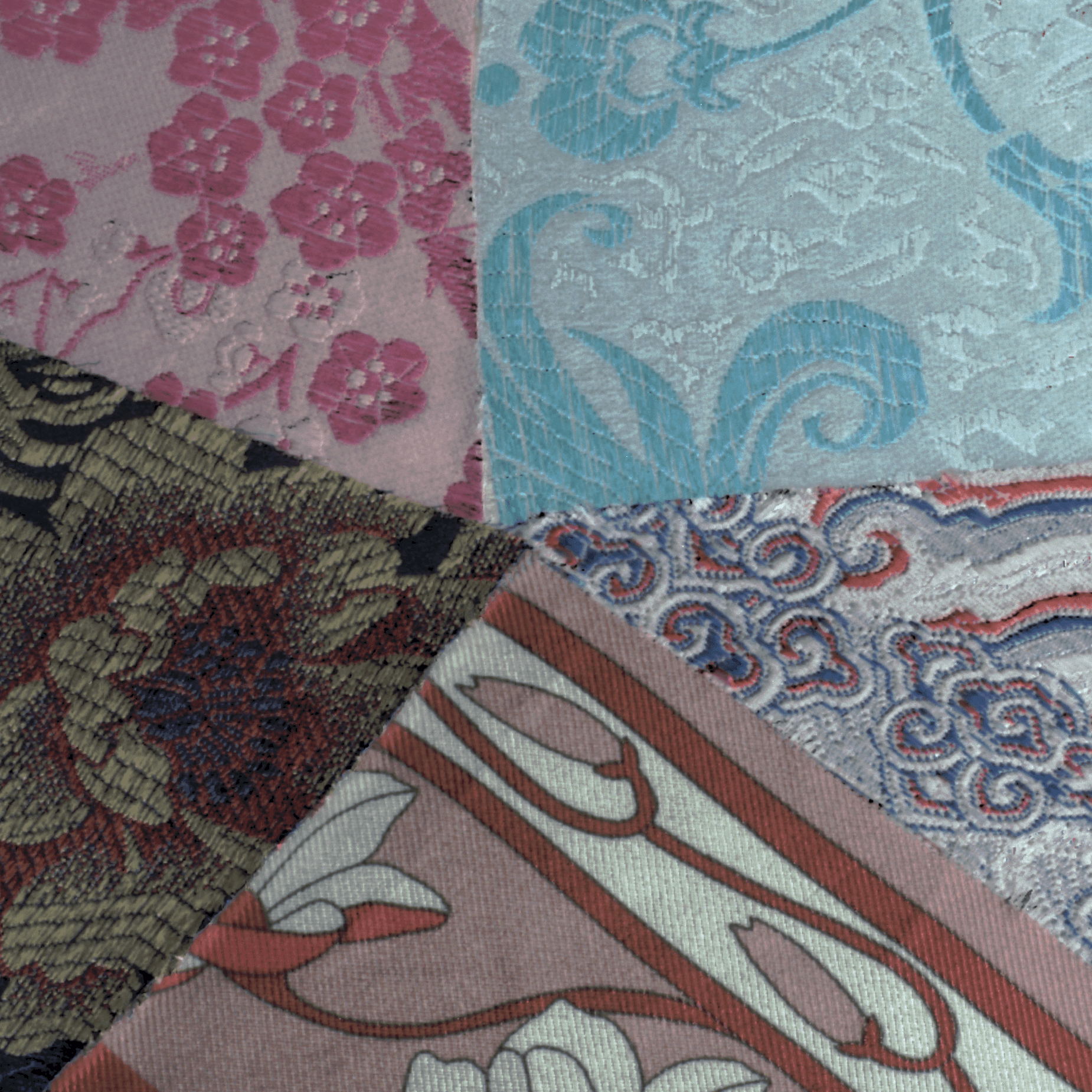}
            \put(-57,3) {\small \color{white} SSIM = 0.85}
        \end{minipage}	

        \begin{minipage}{0.32\linewidth}	
            \includegraphics[width=\linewidth]{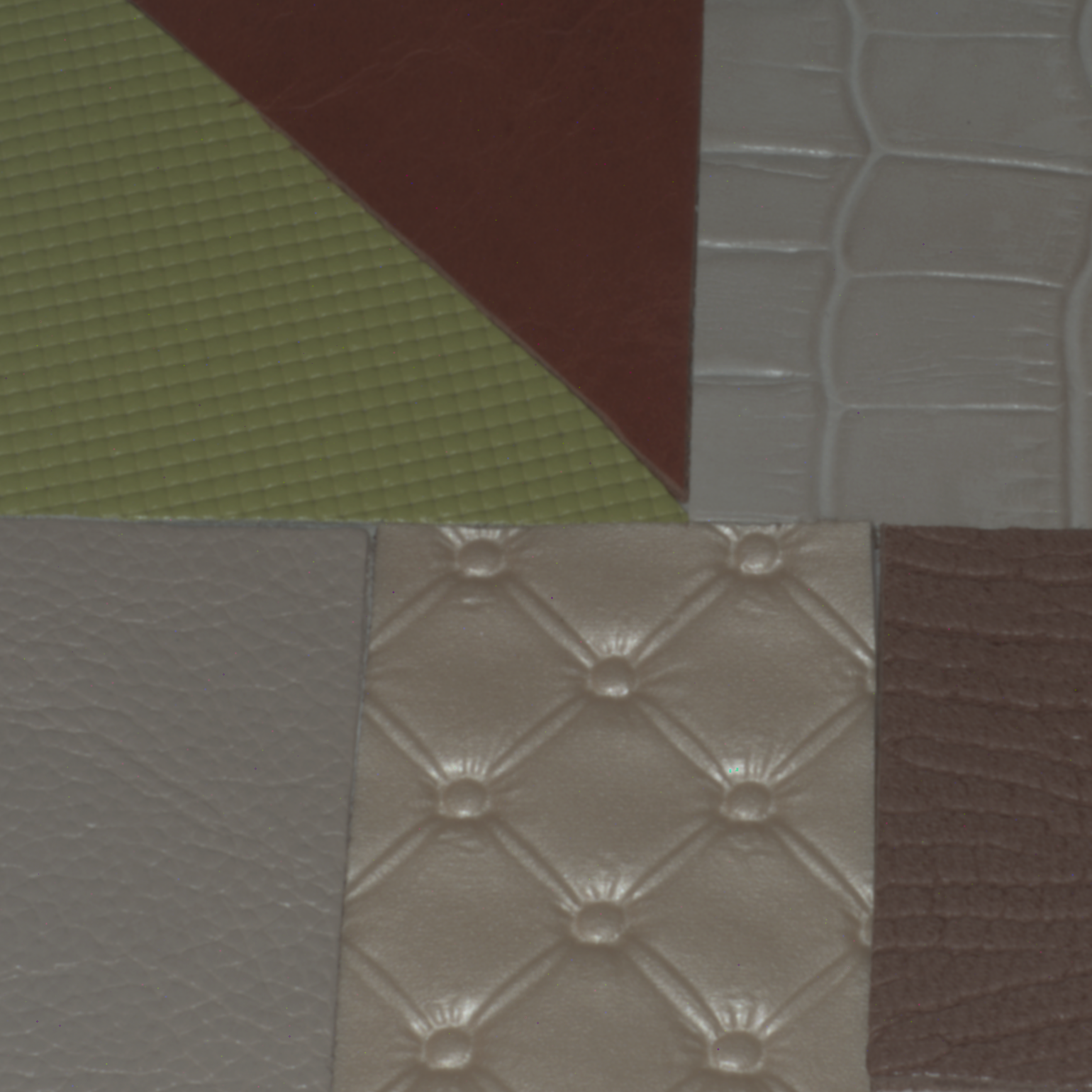}
        \end{minipage}	
        \begin{minipage}{0.32\linewidth}	
            \includegraphics[width=\linewidth]{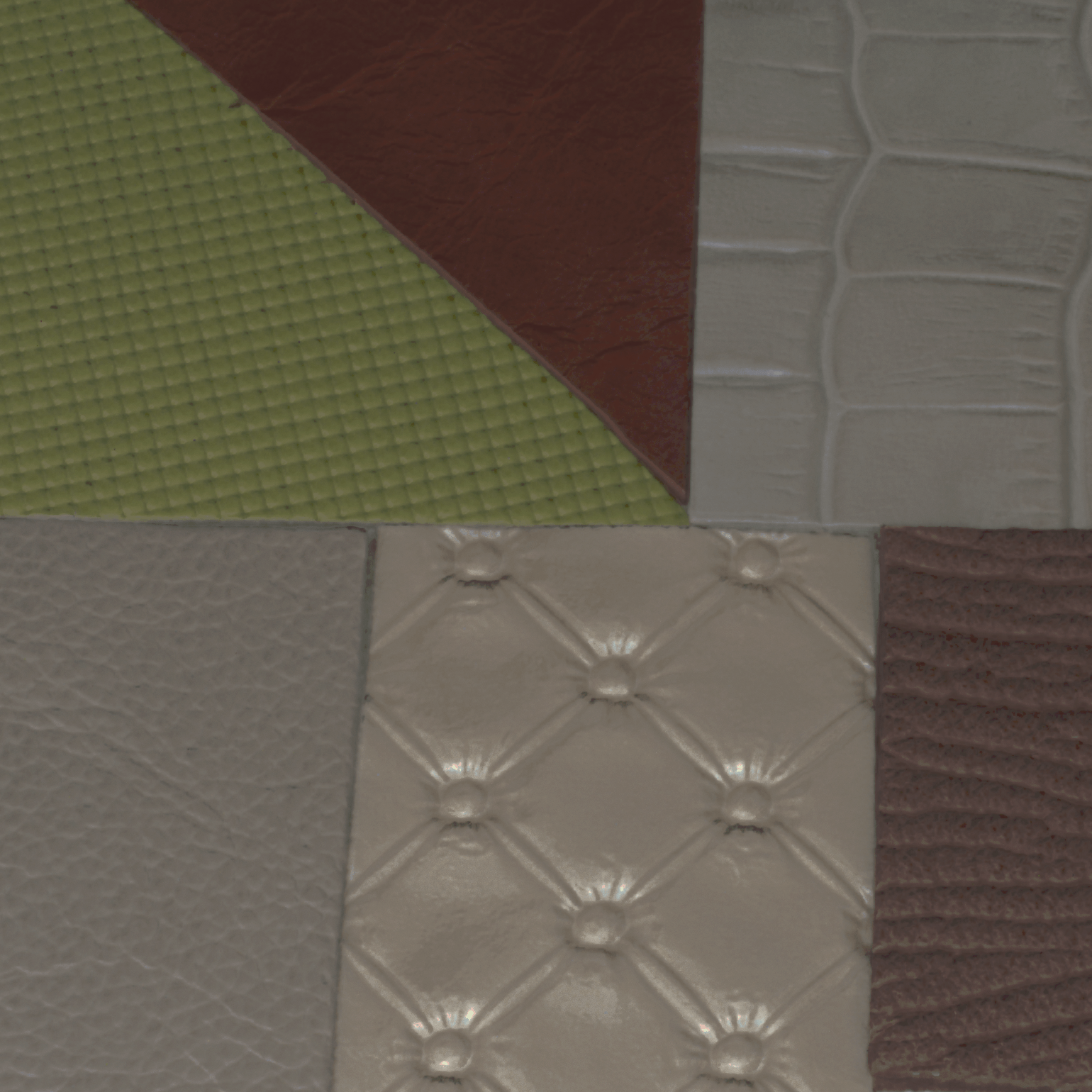}
            \put(-57,3) {\small \color{white} SSIM = 0.95}
        \end{minipage}	
        \begin{minipage}{0.32\linewidth}	
            \includegraphics[width=\linewidth]{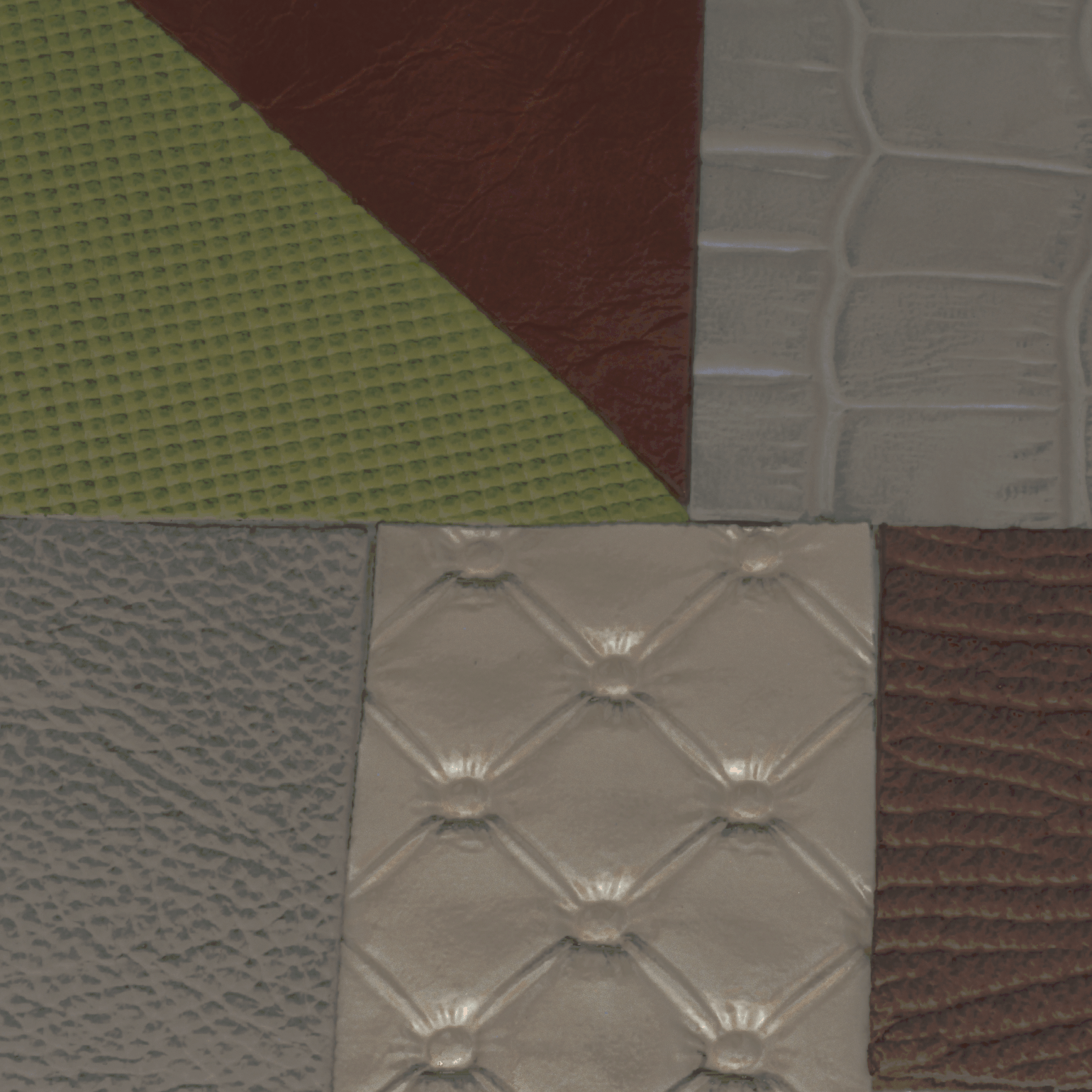}
            \put(-57,3) {\small \color{white} SSIM = 0.92}
        \end{minipage}	

        \begin{minipage}{0.32\linewidth}	
            \includegraphics[width=\linewidth]{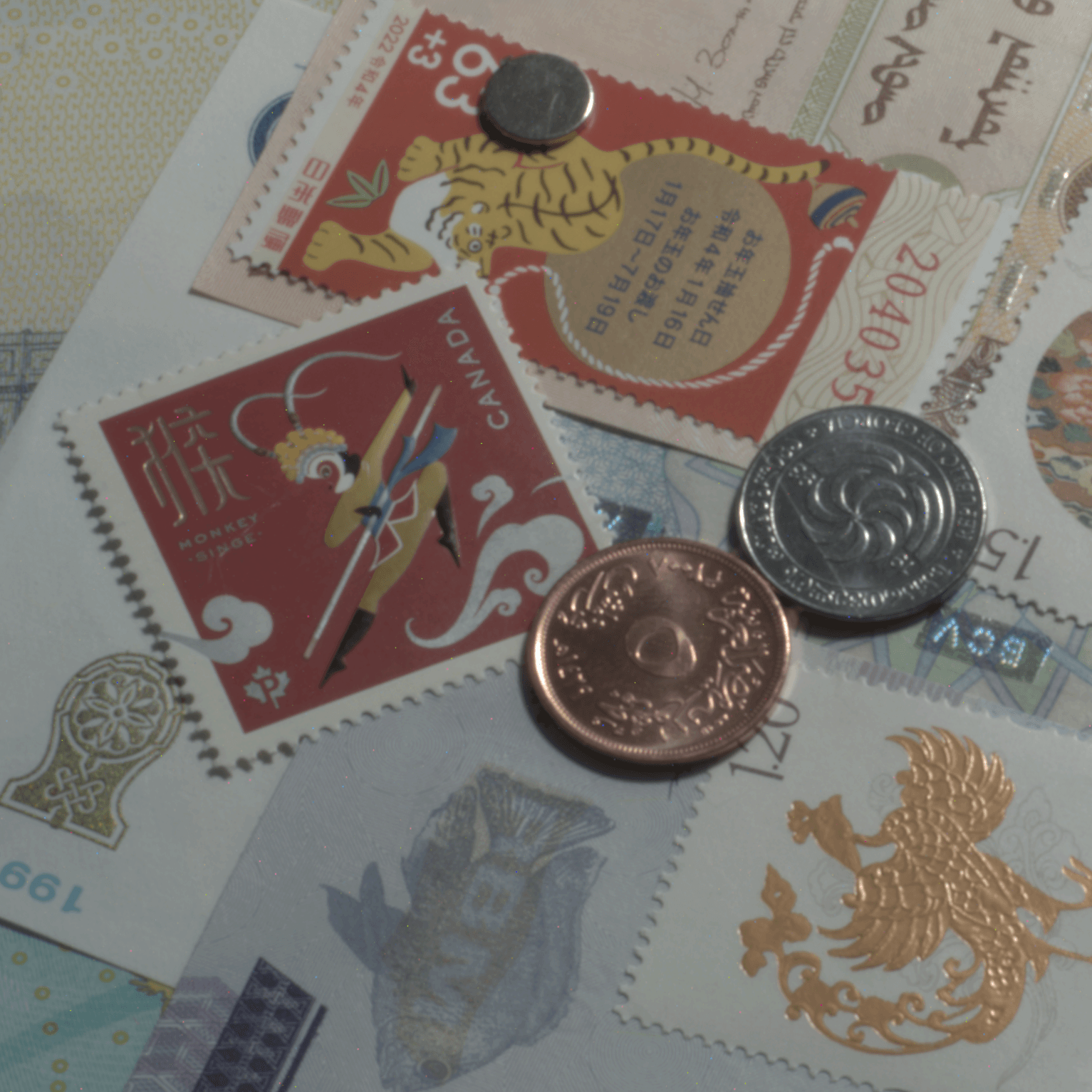}
        \end{minipage}	
        \begin{minipage}{0.32\linewidth}	
            \includegraphics[width=\linewidth]{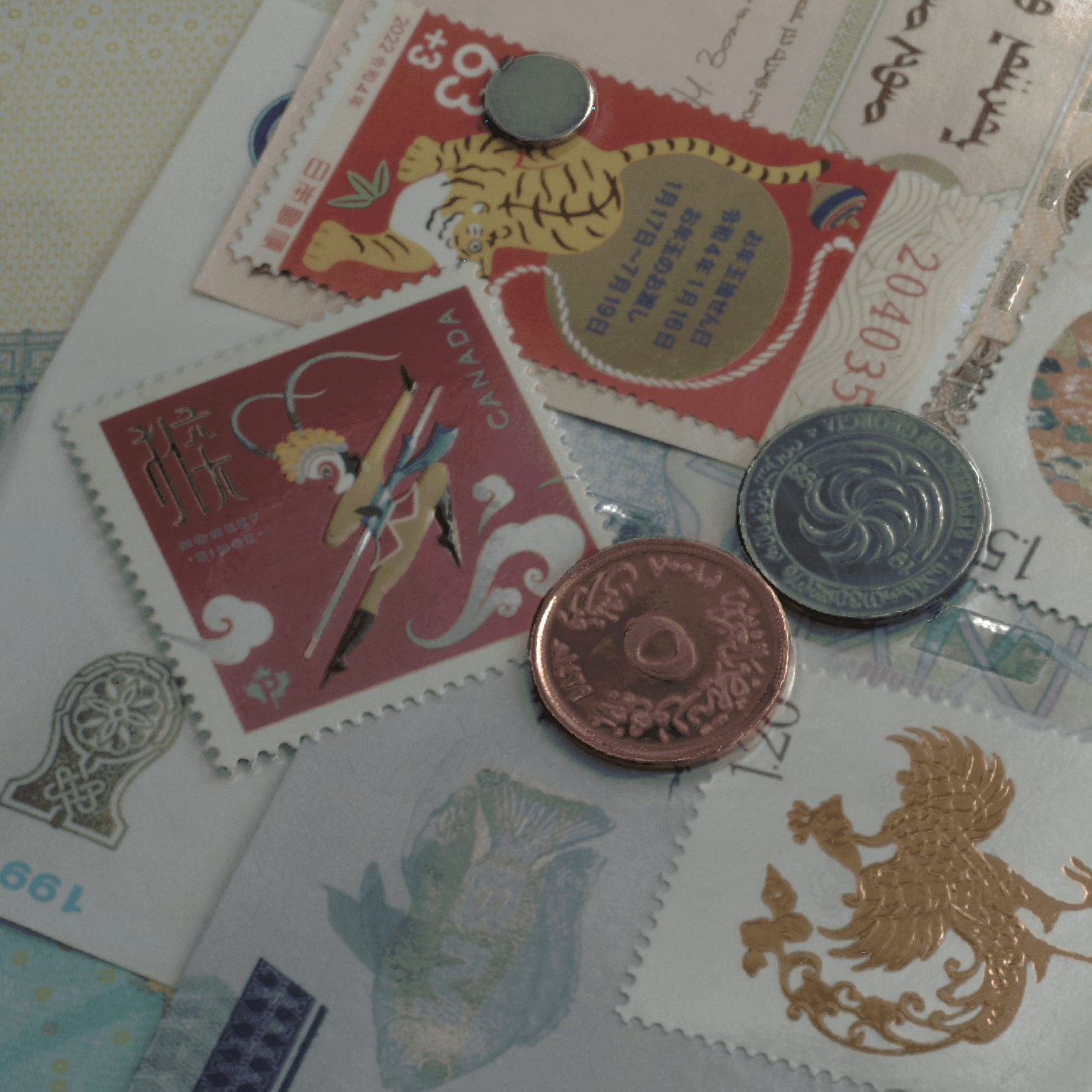}
            \put(-57,3) {\small \color{white} SSIM = 0.94}
        \end{minipage}	
        \begin{minipage}{0.32\linewidth}	
            \includegraphics[width=\linewidth]{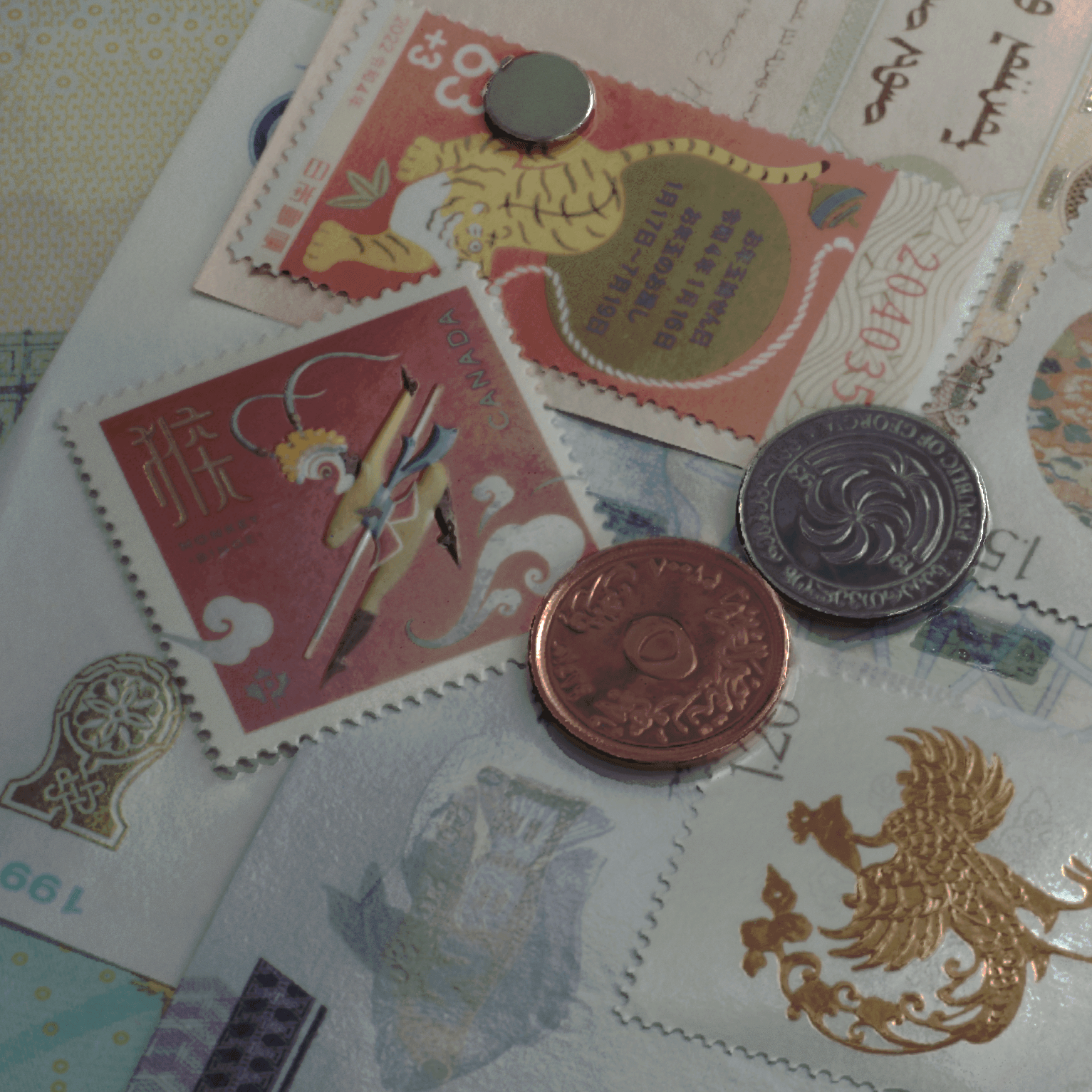}
            \put(-57,3) {\small \color{white} SSIM = 0.92}
        \end{minipage}	
        
        \begin{minipage}{0.32\linewidth}	
            \includegraphics[width=\linewidth]{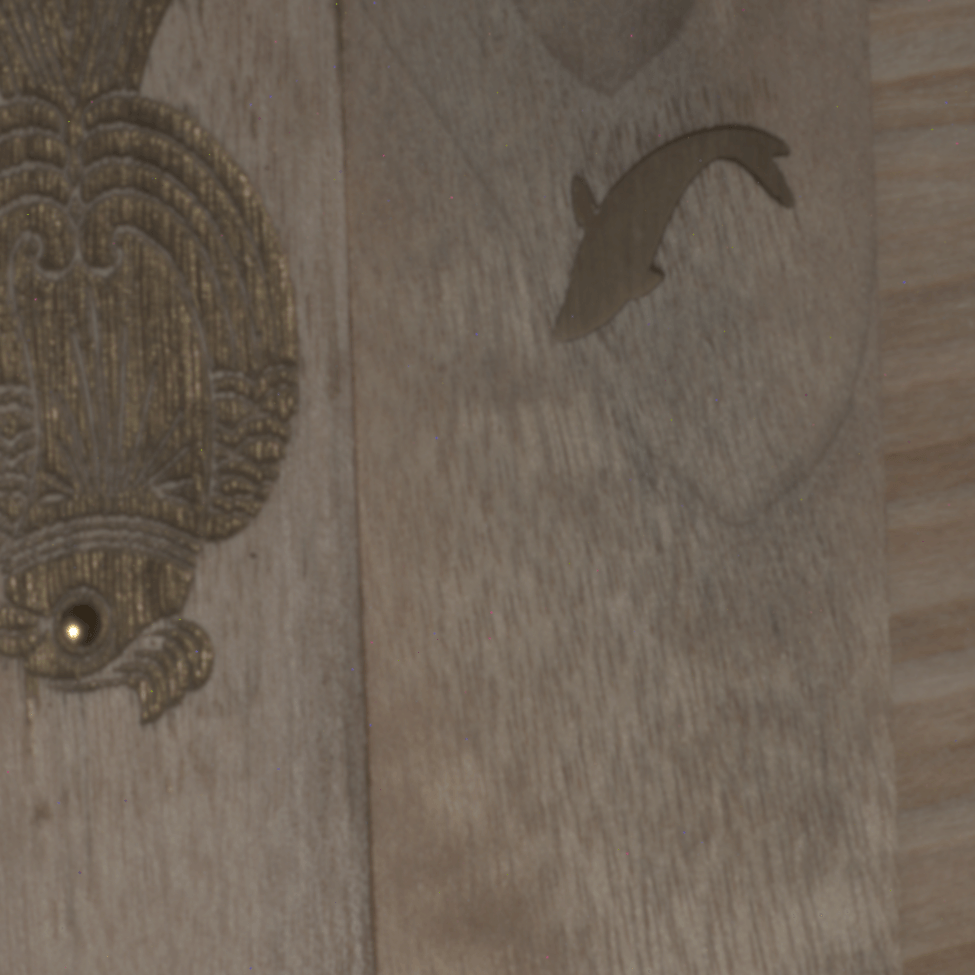}
        \end{minipage}	
        \begin{minipage}{0.32\linewidth}	
            \includegraphics[width=\linewidth]{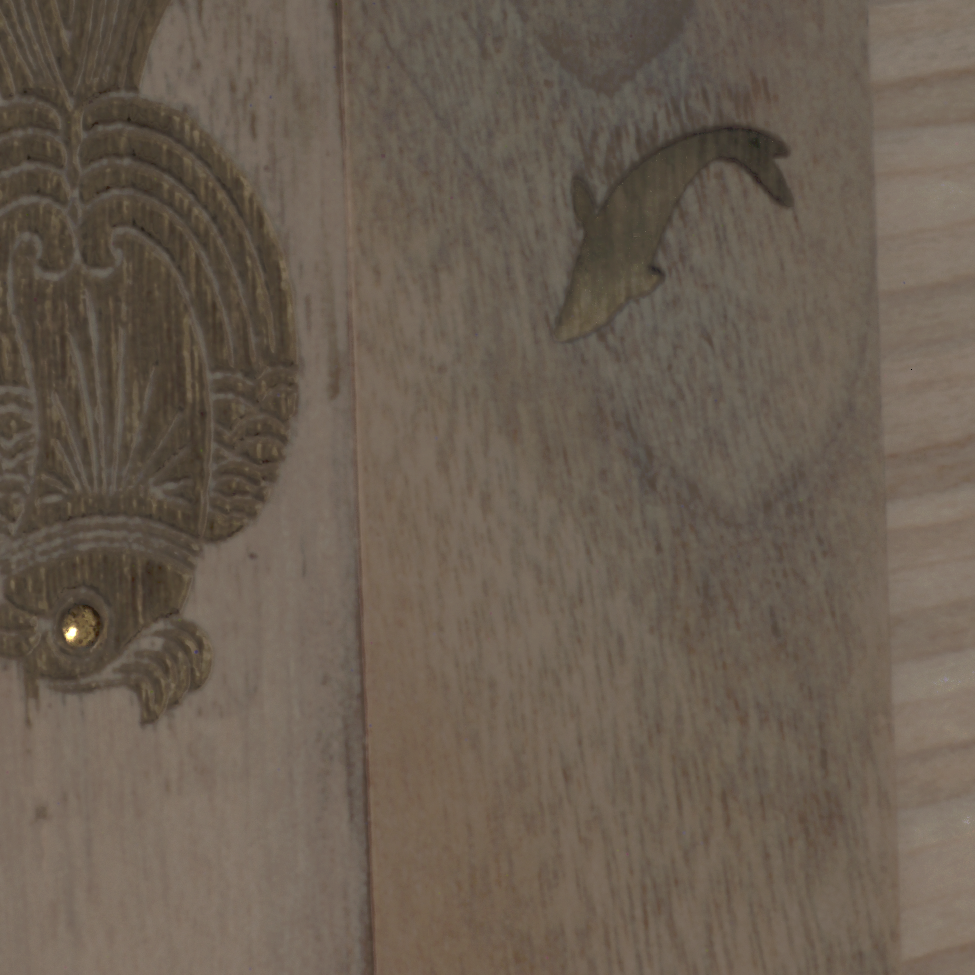}
            \put(-57,3) {\small \color{white} SSIM = 0.97}
        \end{minipage}	
        \begin{minipage}{0.32\linewidth}	
            \includegraphics[width=\linewidth]{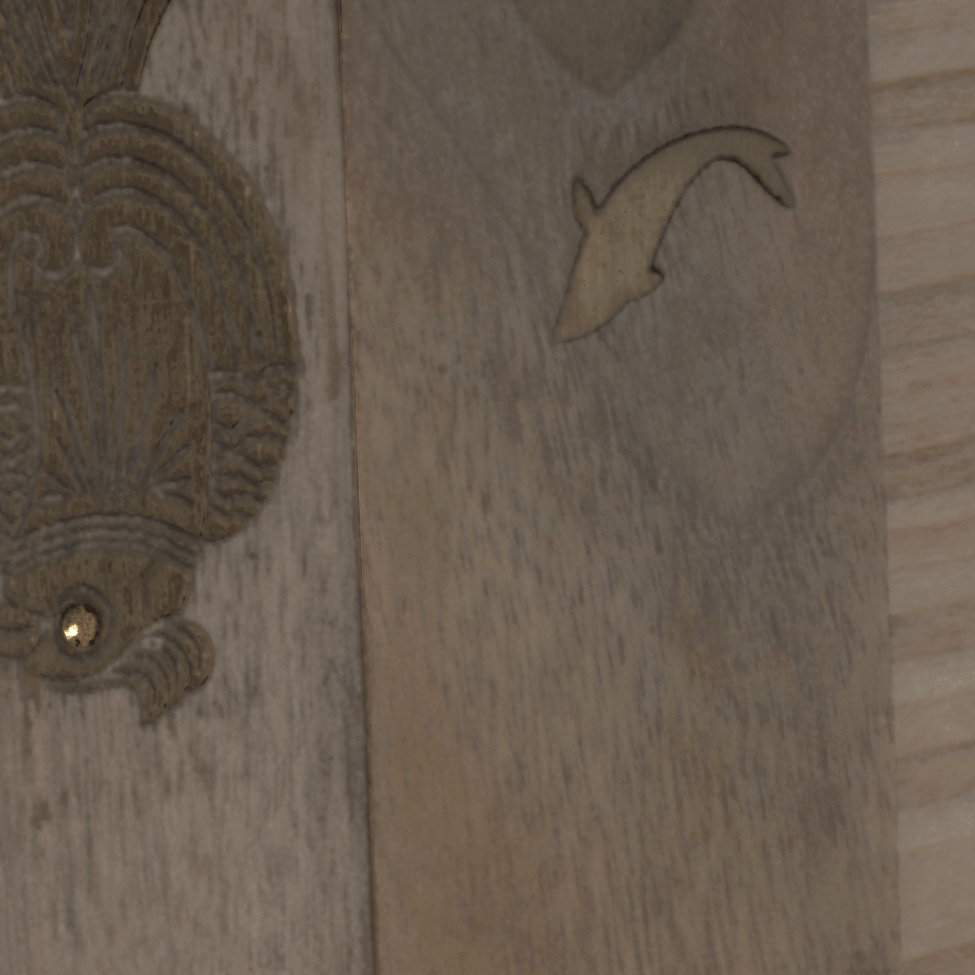}
            \put(-57,3) {\small \color{white} SSIM = 0.95}
        \end{minipage}

     \end{minipage}       

  \caption{Photograph validations. From the left column to right, a photograph of the physical sample set, our result and the result with LDAE. Quantitative errors with respect to the photographs are reported in SSIM on corresponding images.}
  \label{fig:validation}
\end{figure}

We plot the validation losses of different networks with different parameters in~\figref{fig:loss}, representing the average reconstruction quality of lumitexels. The horizontal axis indicates the input bandwidth (i.e., the lighting pattern number $\#$), and the vertical axis shows the network loss $L$ (\eqnref{eq:loss}). For the vanilla version, our network consistently outperforms LDAE at the same bandwidth (as also shown in~\figref{fig:validation}), marked as yellow and red dots. 

Since the size of our network is about twice that of LDAE, we double the capacity of their network and find that the validation loss stays on the same level, marked as LDAE(2x). This demonstrates the benefit of our architecture over LDAE at similar capacities. In addition, we test a variant of our network with only half the decoders. The corresponding loss rises slightly, marked as Ours(64 dec.). Therefore, we suggest that more decoders should be employed to improve reconstruction quality, if time and resource permit. Finally, we switch the lighting patterns in our network to fixed ones, obtained by applying principal component analysis to a large number of synthetic lumitexels, marked as PCA. The loss increases substantially, demonstrating the efficiency in using jointly trained lighting patterns.

\section{Limitations \& Future Work}
Our work shares similar limitations with existing work on neural acquisition~\cite{Kang:2018:AUTO}, including unexpected output on physical lumitexels that substantially deviate from training data, and no considerations for global illumination.

In the future, it will be promising to further improve the acquisition efficiency, by performing additional multiplexing in the spectral domain~\cite{Ma:2021:SCANNER}. We would also like to explore more load-balancing techniques, and apply our architecture to boost other work on neural acquisition. Finally, it will be interesting to extend to handle more general appearance, such as subsurface scattering.

\bibliographystyle{ACM-Reference-Format}
\bibliography{neural-tree.bib}

\appendix

\end{document}